\documentclass[twoside,12pt]{article}

\usepackage{jmlr2e}

\usepackage{setspace}
\usepackage{bm,amsmath}
\usepackage[utf8x]{inputenc}
\usepackage{mathrsfs}
\usepackage{xcolor}

\usepackage{geometry}
\newcommand{\stkout}[1]{\ifmmode\text{\sout{\ensuremath{#1}}}\else\sout{#1}\fi}

\newtheorem{assumption}{Assumption}

\newcommand{\tr}{\mathrm{tr}}
\newcommand{\diag}{\mathrm{diag}}

\ShortHeadings{Robustness of $\alpha$-posteriors and their variational approximations}{Avella Medina and Montiel Olea and Rush and Velez}
\firstpageno{1}

\begin{document}

\title{On the Robustness to Misspecification of $\alpha$-Posteriors and Their Variational  Approximations}

\author{\name Marco Avella Medina \email marco.avella@columbia.edu \\
       \addr Department of Statistics\\
       Columbia University\\
       New York, NY 10027, USA
       \AND
     \name Jos\'{e} Luis Montiel Olea \email jm4474@columbia.edu \\
       \addr Department of Economics\\
       Columbia University\\
       New York, NY 10027, USA
       \AND
        \name Cynthia Rush \email cynthia.rush@columbia.edu \\
       \addr Department of Statistics\\
       Columbia University\\
       New York, NY 10027, USA
       \AND
        \name Amilcar Velez \email amilcare@u.northwestern.edu \\
       \addr Department of Economics\\
       Northwestern University\\
       Evanston, IL 60208, USA
       }
\maketitle

\begin{abstract}
$\alpha$-posteriors and their variational approximations distort standard posterior inference by downweighting the likelihood and introducing variational approximation errors. We show that such distortions, if tuned appropriately, reduce the Kullback-Leibler (KL) divergence from the true, but perhaps infeasible, posterior distribution when there is potential parametric model misspecification. To make this point, we derive a Bernstein-von Mises theorem showing convergence in total variation distance of $\alpha$-posteriors and their variational approximations to limiting Gaussian distributions. We use these distributions to evaluate the KL divergence between true and reported posteriors. We show this divergence is minimized by choosing $\alpha$ strictly smaller than one, assuming there is a vanishingly small probability of model misspecification. The optimized value becomes smaller as the the misspecification becomes more severe. The optimized KL divergence increases logarithmically in the degree of misspecification and not linearly as with the usual posterior.
\end{abstract}


\onehalfspacing

\section{Introduction}
A recent body of work in Bayesian statistics \citep{grunwald2011safe, holmes2017assigning, grunwald2012safe,bhattacharyaetal2019,milleranddunson2019, Knoblauch:2019} and probabilistic machine learning \citep{huang2018improving, higgins2017beta, burgess2018understanding} has analyzed the statistical properties of $\alpha$-posteriors and their variational approximations. The $\alpha$-posteriors---also known as fractional, tempered, or power posteriors---are proportional to the product of the prior and the $\alpha$-power of the  likelihood \citep[Chapter 8.6]{ghosal2017}. Their variational approximations are defined as those distributions within some tractable subclass that minimize the Kullback-Leibler (KL) divergence to the $\alpha$-posterior; see  \cite{alquierandridgway2020}, Definition 1.2.   

We contribute to this growing literature by investigating the robustness-to-misspecification of $\alpha$-posteriors and their variational approximations, with a focus on low-dimensional, parametric models.  
Our analysis---motivated by the seminal work of \cite{Gustafson2001}---is based on a simple idea. Suppose two different procedures lead to \emph{incorrect} a posteriori inference (either due to a likelihood misspecification or computational considerations). Define one procedure to be more robust than the other if it is  closer---in terms of KL divergence---to the \emph{true} posterior. Is it true that $\alpha$-posteriors (or their variational approximations) are more robust than standard Bayesian inference?

We answer this question using asymptotic approximations. We establish a Bernstein-von Mises theorem (BvM) in total variation distance for $\alpha$-posteriors (Theorem \ref{thm:BVMT-alpha}) and for their (Gaussian mean-field) variational approximations (Theorem \ref{thm:BVMT-alpha-var}). Our result allows for both model misspecification and non i.i.d.\ data. The main assumptions are that the likelihood ratio of the presumed model is stochastically locally asymptotically normal (LAN) as in \cite{kleijn2012bernstein} and that the $\alpha$-posterior concentrates around the (pseudo)-true parameter at rate $\sqrt{n}$. Our theorem generalizes the results in   \cite{wang2019variational,WangBleiJASA2019}, who focus on the case in which $\alpha = 1$. We also extend the results of \cite{Yu2019}, who establish the BvM theorem for $\alpha$-posteriors under a weaker norm, but under more primitive conditions.
   
These asymptotic distributions allows us to study our suggested measure of robustness by computing the KL divergence between multivariate Gaussians with parameters that depend on the data, the sample size, and the `curvature' of the likelihood. One interesting observation is that relative to the BvM theorem for the standard posterior or its variational approximation, the choice of $\alpha$ only re-scales the limiting variance, with no mean effect. The new scaling is as if the observed sample size were $\alpha \cdot n$ instead of $n$, but the location for the Gaussian approximation continues to be the maximum likelihood estimator. Thus, the posterior mean of $\alpha$-posteriors and their variational approximations has the same limit regardless of the value of $\alpha$.     

When computing our measures of robustness, we think of a researcher that, ex-ante, places some small exogenous probability $\epsilon_n$ of model misspecification and is thus interested in computing an \emph{expected} KL. This forces us to analyze the KL divergence under both correct and incorrect specification. Under the assumption that as the sample size $n$ increases, the probability of misspecification decreases as $ n \epsilon_n \rightarrow \varepsilon$ for constant $\varepsilon \in (0, \infty)$, we establish three main results (Theorem 3) that we believe speak to the robustness of $\alpha$-posteriors and their variational approximations.

The first result shows that for a large enough sample size, the expected KL divergence between $\alpha$-posteriors and the true posterior is minimized for some $\alpha^*_n \in (0,1)$. This means that, for a properly tuned value of $\alpha \in (0,1)$, inference based on the $\alpha$-posterior is asymptotically more robust than regular posterior inference (corresponding to the case $\alpha =1$). Our calculations suggest that $\alpha^*_n$ decreases as both the probability of misspecification $\epsilon_n$ and the difference between the true and pseudo-true parameter increase, where by pseudo-true parameter we mean the point in the parameter space that provides the best  approximation (in terms of KL divergence) to the true data generating process. In other words, this analysis makes the reasonable suggestion that as the probability of the likelihood function being wrong increases, one should put less emphasis on it when computing the posterior.

The second result demonstrates that the Gaussian mean-field variational approximations of $\alpha$-posteriors inherit some of the robustness properties of $\alpha$-posteriors. In particular, it is shown that the expected KL divergence between the true posterior and the \emph{mean-field} variational approximation to the $\alpha$-posterior is also minimized at some $\widetilde{\alpha}^{*}_n \in (0,1)$. Our second result thus provides support to the claim that variational approximations to $\alpha$-posteriors are more robust to model misspecification than regular variational approximations (corresponding to $\alpha =1$). Our result also provides some theoretical support for the recent work of \cite{higgins2017beta} that suggests introducing an additional hyperparameter $\beta$ to temper the posterior when implementing variational inference (see Eq.\ \ref{eq:new_optimization} and the discussion that follows it).

The final result contrasts the expected KL of the optimized $\alpha$-posterior (and also of the optimized $\alpha$-variational approximation) against the expected KL of the regular posterior. We find that the latter increases linearly in the magnitude of misspecification (measured by the difference between the true parameter $\theta_0$ and the pseudo-true parameter $\theta^*$), while the former does so logarithmically. This suggests that when the model misspecification is large, there will be significant gains in robustness from using $\alpha$-posteriors and their variational approximations.  

\subsection{Related Work}
The idea that Bayesian procedures can be enhanced by adding an $\alpha$ parameter to decrease the influence of the likelihood in the posterior calculation has been suggested independently by many authors in the past, predating the more recent research studying its robustness properties under the name of $\alpha$-posteriors. This includes work by \cite{vovk1990aggregating}, \cite{mcallester2003pac}, \cite{barron1991minimum}, \cite{walker2001bayesian}, and \cite{zhang2006e}.

A large part of the theoretical literature studying the robustness of $\alpha$-posteriors and their variational approximations has focused on nonparametric or high-dimensional models. In these works, the term robustness has been used to mean that, in large samples,  $\alpha$-posteriors and their variational approximations \emph{concentrate} around the pseudo-true parameter, even when the standard posterior does not. \cite{bhattacharyaetal2019} illustrate this point by providing examples of heavy-tailed priors in both density estimation and regression analysis where the $\alpha$-posterior can be guaranteed to concentrate at (near) minimax rate, but the regular posterior cannot.  \cite{alquierandridgway2020} also derive concentration rates for $\alpha$-posteriors for high-dimensional and non-parametric models. Although comparison of the conditions required to verify concentration properties is natural in nonparametric or high-dimensional models, for the analysis of low-dimensional parametric models there are alternative suggestions in the literature. For example, \cite{wang2019variational} compare the posterior predictive distribution based on the regular posterior and also its variational approximation, and they show that under likelihood misspecification, the difference between the two posterior predictive distributions converges to zero. \cite{grunwald2017inconsistency} also used (in-sample) predictive distributions to assess the performance of $\alpha$-posteriors. They show that in a misspecified linear regression problem (where the researcher assumes homoskedasticity, but the data is heteorskedastic) $\alpha$-posteriors can outperform regular posteriors.  

Our work complements this recent body of work by demonstrating robustness in a different sense, while still connecting to previous work on the subject. While it is true that one may  be content with just studying `first order' properties of $\alpha$-posteriors (like contraction) whenever the BvM does not hold, it is not yet clear how to properly formalize the first order benefits of $\alpha$-posteriors and their variational approximations under misspecification. \cite{yang2020alpha} show that the necessary conditions for optimal contraction in misspecified models can be relaxed, but there are no results yet that formally tease apart the benefits of $\alpha$-posteriors compared to the usual posteriors.
Our results exploit heavily the BvM theorem which, in a sense, is based on `second order' approximations.  We believe that our results may be useful in studying other models where BvM results are available, but are more complicated than that considered in this paper, in that the underlying model is not necessarily low-dimensional and/or parametric; for example, semiparametric models where the object of interest are smooth functionals as in \cite{castillo2015bernstein}. %This is, however, left for future work.

More importantly, our results are an attempt to provide an additional rationale for the use of variational inference methods due to their robustness to model misspecification, as opposed to solely based on computational considerations. We hope that our results will encourage the use of variational methods in fields like applied statistics and econometrics where variational inference remains somewhat underutilized despite its tremendous impact in machine learning. Recent applications of variational inference in econometrics include \cite{bonhomme2021teams}, \cite{mele2019approximate}, and \cite{koop2018variational}.   

\subsection{Organization}
The rest of this paper is organized as follows. Section \ref{section:Notation} presents definitions, notations, and our general framework. Section \ref{section:BvM} presents the Bernstein-von Mises theorem for $\alpha$-posteriors and their variational approximations. Section \ref{section:Robustness} presents our main results on the theoretical analysis of our suggested measure of robustness. 
Section \ref{section:Example} presents an example concerning a Gaussian linear regression model with omitted variables. 
Section \ref{section:Conclusion} presents some concluding remarks and discussion. The proofs of our main results are collected in Appendix \ref{section:Main_Appendix}.

\section{Notation General Framework}
\label{section:Notation}

\subsection{Paper Notation and Definitions}
We begin by introducing notation and definitions that we will use throughout the paper.
We use $\phi(\cdot | \mu, \Sigma)$ to denote the density of a multivariate normal random vector with mean $\mu$ and covariance matrix $\Sigma$. The indicator function of event $A$, namely the function that takes the value $1$ if event $A$ occurs and $0$ otherwise is denoted $\mathbf{1}\{ A \}$. If $p$ and $q$ are two densities with respect to Lebesgue measure in $\mathbb{R}^{p}$, the total variation distance between them, denoted $d_{\textrm{TV}}(p, \, q)$, equals
\begin{equation} \label{eqn:TV}
    d_{\textrm{TV}}(p, \, q) \equiv \frac{1}{2} \int_{\mathbb{R}^{p}}  \, |\, p(u)-q(u) \, | \, du \, .
\end{equation}
See Section 2.9 in \cite{van2000asymptotic} for more details on the total variation distance.
The KL divergence between the two distributions $p$ and $q$, denoted $\mathcal{K}(p \: || \:  q)$, is defined as
\begin{equation}
\mathcal{K}( p \: ||\:  q )  \equiv \int p(u) \log \left(\frac{p(u)}{ q(u)} \right) d u \, .
\label{eq:KL}
\end{equation}
For a sequence of densities $p_n$ on random variables $X_n$, we say that $X_n = o_{p_{n}}(1)$ if $\lim_n \mathbb{P}_{p_{n}}(\|X_n\| > \epsilon) = 0$ for every $\epsilon > 0$ and that the sequence $X_n$ is `bounded in $p_{n}$-probability' if for every $\epsilon>0$ there exists $M_{\epsilon}>0$ such that $\mathbb{P}_{p_{n}}(\|X_n\|<M_{\epsilon}) \geq 1- \epsilon$.

\subsection{Statistical Model}

Let $\mathcal{F}_n \equiv \{f_n(\cdot \, | \, \theta): \theta \in \Theta \subseteq \mathbb{R}^{p} \}$ be a parametric family of densities used as a statistical model for the random vector $X^n \equiv (X_1, \ldots, X_n)$. The statistical model may be  \emph{misspecified}, in the sense that if $f_{0,n}$ is the true density for the random vector $X^n$, then $f_{0,n}$ may not belong to $\mathcal{F}_n$. As usual, we define the maximum likelihood (ML) estimator---denoted by  $\widehat{\theta}_{\textrm{ML\,-\,$\mathcal{F}_n$}}$---as
\begin{equation} \label{eqn:ML}
 \widehat{\theta}_{\textrm{ML\,-\,$\mathcal{F}_n$}} \equiv \underset{\theta \, \in \, \Theta}{\arg \max} \, f_n (X^n \, |  \, \theta) \,  .
\end{equation}
For simplicity, we assume that the ML estimator is unique, and that there is a parameter value $\theta^*$ in the interior of $\Theta$ for which 
$\sqrt{n} \,(\widehat{\theta}_{\textrm{ML\,-\,$\mathcal{F}_n$}}-\theta^*)$ is asymptotically normal. Sufficient conditions for the consistency and asymptotic normality of the ML estimator under model misspecification with i.i.d.\ data can be found in \cite{huber1967behavior,white1982,kleijn2012bernstein}. Examples of other papers establishing consistency and asymptotic normality under misspecification for certain types of non i.i.d.\ data are given in \cite{pouzo2016maximum}. 
If the model is correctly specified, then $\theta^*$ is simply the true parameter, but if the model is misspecified, then $\theta^*$ provides the best  approximation (in terms of KL divergence) to the true data generating process. In the latter misspecified case, it is then common to refer to $\theta^*$ as the \emph{pseudo-true} parameter. We focus on the case in which the ML estimator is asymptotically normal to highlight the fact that none of our results depend on atypical asymptotic distributions. 

The main restriction we impose on $\mathcal{F}_n$ is that the $\sqrt{n}$-stochastic local asymptotic normality (LAN) condition of \cite{kleijn2012bernstein}, around $\theta^*$. 
\begin{assumption} \label{asn:A1}
Denote $\Delta_{n,\theta^*} \equiv \sqrt{n} \, (\widehat{\theta}_{\textrm{ML\,-\,$\mathcal{F}_n$}}-\theta^*)$. There exists a positive definite matrix $V_{\theta^*}$ such that 
\begin{equation}
    R_n(h) \equiv \log \left( \frac{f_n\left(X^n \, | \, \theta^* + h/\sqrt{n}\right)}{f_n \left(X^n \, |  \, \theta^*\right )}  \right) - h^{\top} \, V_{\theta^*} \, \Delta_{n,\theta^*}  + \frac{1}{2} \, h^{\top} \, V_{\theta^*}\, h \, ,
\end{equation}
satisfies 
\begin{equation}
    \sup_{h \in K} \, \left |  R_n(h)  \right| \rightarrow 0 \,  ,
\end{equation}
in $f_{0,n}$-probability, for any compact set $K \subseteq \mathbb{R}^p$. 
\end{assumption}
Noticing that 
\begin{equation}
    h^{\top} \, V_{\theta^*} \, \Delta_{n,\theta^*}  - \frac{1}{2} \, h^{\top} \, V_{\theta^*}\, h = 
    \log \left(\frac{\phi(\theta^* + h/\sqrt{n} \, | \, \widehat{\theta}_{\textrm{ML\,-\,$\mathcal{F}_n$}}, \, (nV_{\theta^*})^{-1})}{\phi(\theta^* \,| \, \widehat{\theta}_{\textrm{ML\,-\,$\mathcal{F}_n$}}, (nV_{\theta^*})^{-1})}\right) \, ,
\end{equation}
we can see that Assumption~\ref{asn:A1} implies that the corresponding likelihood ratio process for $\mathcal{F}_n$ approximates, asymptotically, that of a normal random variable.

\subsection{$\alpha$-posteriors and their Variational Approximations}
We now present the definition of $\alpha$-posteriors and their variational approximations. Starting from the statistical model $\mathcal{F}_n$, a prior density $\pi$ for $\theta$, and a scalar $\alpha>0$, the $\alpha$-posterior is defined as the distribution having density:
\begin{equation} \label{eqn:alpha-posterior}
    \pi_{n,\alpha}(\theta \, | \, X^n) \equiv  \frac{\left [ f_n(X^n \, | \, \theta) \right] ^\alpha \pi(\theta)}{ \int \left[  f_n(X^n \, | \,  \theta) \right] ^\alpha \pi(\theta)  \, d\theta  } \, .
\end{equation}
See Chapter 8.6 in \cite{ghosal2017} for a textbook definition.

The projection of \eqref{eqn:alpha-posterior}---in Kullback-Leibler (KL) divergence---onto the space of probability distributions with independent marginals, also referred to as the mean-field family and denoted $\mathcal{Q}_{\textrm{MF}}$, provides the \emph{mean-field variational approximation} to the $\alpha$-posterior:
\begin{equation}\label{eqn:mf-alpha-posterior}
    \widetilde{\pi}_{n,\alpha}(\cdot \, | \,  X^n) \in \underset{q \, \in \,  \mathcal{Q}_{\textrm{MF}}}{\arg \min} \: \mathcal{K} ( q \: || \: \pi_{n,\alpha}(\cdot \, | \, X^n)  ) \, . 
\end{equation}

There is a trade-off between choosing a flexible and rich enough domain for the optimization in \eqref{eqn:mf-alpha-posterior}, so that $q$ can be close to $\pi_{n,\alpha}(\theta  \,| \, X^n)$, and choosing a domain that is also constrained enough such that the optimization is computationally feasible. The projection onto the Gaussian mean-field family, studied in equation \eqref{eqn:mf-alpha-posterior}, is a particular case of the more general variational approximations studied in the recent work of   \cite{alquierandridgway2020}, who allow for other sets of distributions over which the KL is minimized. A key insight of the variational framework is that minimizing the KL divergence in \eqref{eqn:mf-alpha-posterior} is equivalent to solving the program
\begin{equation}
 \widetilde{\pi}_{n,\alpha}(\cdot \, | \,  X^n) \equiv \underset{q \, \in \,  \mathcal{Q}_{\textrm{MF}}}{\arg \min} \: \left \{\int q(\theta) \log \left( f_n(X^n \, | \,  \theta) \right) \, d \theta  - (1/\alpha) \, \mathcal{K}( q \: || \: \pi )  \right\} \, .
\label{eq:new_optimization}
\end{equation}
The objective function in \eqref{eq:new_optimization} is reminiscent of penalized estimation: it involves a data-fitting term (the average log-likelihood) and a regularization or penalization term that forces the distribution $q$ to be close to a baseline prior $\pi$ with regularization parameter $1/\alpha$.  

The optimization scheme in \eqref{eq:new_optimization} %(with KL divergence)
has been the subject of recent work in the representation learning literature in \cite{burgess2018understanding} and \cite{higgins2017beta} under the name of the $\beta$-variational autoencoder.  The optimization problem has also been studied in axiomatic decision theory; see, for example, the \emph{multiplier preferences} introduced in \cite{hansen2001robust} and their axiomatization in \cite{strzalecki2011axiomatic}. More generally, the objective function  in \eqref{eq:new_optimization}  with an arbitrary divergence function is analogous to the so-called \emph{divergence preferences} studied in \cite{maccheroni2006ambiguity}. We think this literature could be potentially useful in understanding the role of the different divergence functions in penalizations, as well as the multiplier parameter $\alpha$.

\section{Bernstein-von Mises Theorem for $\alpha$-posteriors and their Variational Approximations}
\label{section:BvM}
Before presenting a formal definition of the measure of robustness used in this paper, we show that $\alpha$-posteriors and their variational approximations are asymptotically normal. This extends the Bernstein-von Mises (BvM) theorem for misspecified models in \cite{kleijn2012bernstein}, which shows that posteriors under misspecified models are asymptotically normal, and the recent Variational BvM of \cite{wang2019variational}, which shows that variational approximations of true posteriors are asymptotically normal. We show that the total variation distance between the studied distribution and its limiting Gaussian distribution converges in probability to zero with growing sample size.

\subsection{BvM for $\alpha$-posteriors}
We say that the $\alpha$-posterior, $\pi_{n,\alpha}(\cdot \, | \, X^n)$ defined in \eqref{eqn:alpha-posterior}, concentrates at rate $\sqrt{n}$ around $\theta^*$ if for every sequence of constants $r_n \rightarrow \infty$,
\begin{equation} \label{eqn:concentration}
    \mathbb{E}_{f_{0,n}} \left[ \,  \int_{} \, \mathbf{1}\left \{ \: \| \sqrt{n} \, (\theta-\theta^*) \| > r_n \: \right \} \, \pi_{n,\alpha} (\theta \, | \, X^n) \, d \theta  \right] \rightarrow 0 \, .
\end{equation}
Notice that
\begin{equation*} 
\begin{split}
        \mathbb{E}_{f_{0,n}}& \left[ \,  \int_{} \, \mathbf{1}\left \{ \: \| \sqrt{n} \, (\theta-\theta^*) \| > r_n \: \right \} \, \pi_{n,\alpha} (\theta \, | \, X^n) \, d \theta  \right] = \mathbb{E}_{f_{0,n}} \left[ \, \mathbb{P}_{\pi_{n,\alpha} (\cdot \, | \, X^n)} \left( \| \sqrt{n} \, (\theta-\theta^*) \| > r_n   \right) \right] \, .
\end{split}
\end{equation*}

\begin{theorem} \label{thm:BVMT-alpha}
Suppose that the prior density $\pi$ is continuous and positive on a neighborhood around the (pseudo-) true parameter $\theta^*$, and that   $\pi_{n,\alpha}(\cdot \, | \, X^n)$ concentrates at rate $\sqrt{n}$ around $\theta^*$, as in \eqref{eqn:concentration}. If Assumption \ref{asn:A1} holds, then
\begin{equation}\label{eqn:Thm1}
     d_{\textrm{TV}}\left( \pi_{n,\alpha}(\cdot \, | \, X^n), \, \phi(\cdot \, | \, \widehat{\theta}_{\textrm{ML\,-\,$\mathcal{F}_n$}}, \,  V^{-1}_{\theta^*} / (\alpha n)  )   \right) \rightarrow 0 \,, 
\end{equation}
in $f_{0,n}$-probability, where $d_{\textrm{TV}}(\cdot, \cdot)$ denotes the total variation distance %between two densities $p$ and $q$ 
and is defined in \eqref{eqn:TV} and $V_{\theta^*}$ is the positive definite matrix satisfying Assumption \ref{asn:A1}.
\end{theorem}
In a nutshell, the theorem states that the $\alpha$-posterior distribution behaves asymptotically as a multivariate normal distribution, centered at the ML estimator, $\widehat{\theta}_{\textrm{ML\,-\,$\mathcal{F}_n$}}$, which is based on the potentially misspecified model $\mathcal{F}_n$. Thus, the theorem shows that the choice of $\alpha$ does not asymptotically affect the location of the $\alpha$-posterior distribution.
However, Theorem \ref{asn:A1} shows that the asymptotic covariance matrix of the $\alpha$-posterior is given by $V_{\theta^*}^{-1}/(\alpha  n)$, hence, the parameter $\alpha$ 
inflates the asymptotic variance when $\alpha<1$, and deflates it otherwise.  The matrix $V_{\theta^*}$ is the second-order term in the stochastic LAN approximation in Assumption \ref{asn:A1}, and its inverse is the usual variance in the BvM theorem for correctly or incorrectly specified models. Intuitively, $V_{\theta^*}$ can be thought of as measuring the curvature of the likelihood.

The proof and its details are presented in Appendix \ref{proof:thm1}. For the sake of exposition, we present a brief intuitive argument for why the result should hold. By assumption, the $\alpha$-posterior concentrates around $\theta^*$ at rate $\sqrt{n}$, in the sense of \eqref{eqn:concentration}. Consider the log-likelihood ratio, for some vector $h \in \mathbb{R}^d$,
\begin{equation} \label{eqn:example_LR}
\log \left( \frac{\pi_{n,\alpha}(\theta^* + h/\sqrt{n} \: | \: X^n)}{\pi_{n,\alpha}(\theta^* \: | \: X^n)} \right) = \log \left( \left[ \frac{f(X^n \: | \: \theta^* + h/\sqrt{n})}{f(X^n \: | \: \theta^* ) }\right]^\alpha \frac{\pi(\theta^*+ h /\sqrt{n})}{\pi(\theta^*)}  \right) \, .
\end{equation}
If $\pi_{n,\alpha}(\cdot \, | \, X^n)$ were exactly a multivariate normal having mean  $\widehat{\theta}_{\textrm{ML\,-\,$\mathcal{F}_n$}}$ and covariance matrix $V_{\theta^*}^{-1}/(\alpha  n)$, the log-likelihood ratio in \eqref{eqn:example_LR} would equal 
\begin{equation} \label{eqn:example_LR_N}
\begin{split}
   h^{\top} (\alpha V_{\theta^*}) \sqrt{n}\, (\widehat{\theta}_{\textrm{ML\,-\,$\mathcal{F}_n$}}-\theta^*) - \frac{1}{2} h^{\top} (\alpha V_{\theta^*})h \, .  
    \end{split}
\end{equation}
The log-likelihood ratios in $\eqref{eqn:example_LR}$ and \eqref{eqn:example_LR_N} are not equal, since $\pi_{n,\alpha}(\cdot \, | \, X^n)$ is not exactly multivariate normal, but the continuity of $\pi$ at $\theta^*$ and the stochastic LAN property of Assumption \ref{asn:A1}  makes them equal up to an $o_{f_{0,n}}(1)$ term.

Most of the work in the proof of Theorem~\ref{thm:BVMT-alpha} consists of relating the closeness in log-likelihood ratios to closeness in total variation distance. The arguments we use to make this connection follow verbatim the arguments used by \cite{kleijn2012bernstein}. To the best of our knowledge the result in Theorem \ref{thm:BVMT-alpha} has not appeared previously in the literature, although Section 4.1 in \cite{Yu2019} present a different version of this result in a weaker metric (convergence in distribution, as opposed to total variation), but using lower-level conditions (as opposed to our high-level assumptions). 

\subsection{BvM for Variational Approximations of $\alpha$-posteriors}

Given that the $\pi_{n,\alpha}$ is close to a multivariate normal distribution, it is natural to conjecture that the variational approximation $\widetilde{\pi}_{n,\alpha}$ will converge to the projection of such multivariate normal distribution onto the mean-field family $\mathcal{Q}_{\textrm{MF}}$. 

To formalize this argument, let $\mathcal{Q}_{\textrm{GMF}-p}$ denote the family of multivariate normal distributions of dimension $p$ with independent marginals. An element in this family is parameterized by a vector $\mu \in \mathbb{R}^{p}$ and a positive semi-definite diagonal covariance matrix $\Sigma \in \mathbb{R}^{p \times p}$. When convenient, we denote such an element as $q(\cdot | \mu, \Sigma)$   and will always implicitly assume that $\Sigma$ is diagonal.

We focus on the \emph{Gaussian mean-field approximation to the $\alpha$-posterior}. Given a sample of size $n$, this can be defined as the  Gaussian distribution with parameters $\widetilde{\mu}_n, \widetilde{\Sigma}_n$ satisfying
\begin{equation} \label{eqn:GMF-approx}
    \widetilde{\pi}_{n,\alpha}(\cdot\,|\, X^n) = q(\cdot\,|\, \widetilde{\mu}_n,\:\widetilde{\Sigma}_n) \in \underset{q \, \in \,  \mathcal{Q}_{\textrm{GMF-p}}}{\arg \min} \: \mathcal{K} ( q (\cdot) \: || \: \pi_{n,\alpha}(\cdot \, | \, X^n)  ) \, .
\end{equation}
Theorem \ref{thm:BVMT-alpha} has shown that $\pi_{n,\alpha}(\cdot \, | \, X^n)$ is close to a  multivariate normal distribution with mean $ \widehat{\theta}_{\textrm{ML\,-\,$\mathcal{F}_n$}}$ and variance  $V^{-1}_{\theta^*} / (\alpha n)$. Let $q(\cdot \, | \, \mu^*_n, \Sigma^*_n)$ denote the multivariate normal distribution in the Gaussian mean-field family closest to such limit. That is, 
\begin{equation} \label{eqn:GMF-approx-limit}
q(\cdot \, | \, \mu^*_n, \Sigma^*_n) \, = \underset{q \, \in \,  \mathcal{Q}_{\textrm{GMF-p}}}{\arg \min} \: \mathcal{K} ( q(\cdot)  \: || \:  \phi(\cdot \, | \, \widehat{\theta}_{\textrm{ML\,-\,$\mathcal{F}_n$}}, \, V_{\theta^*}^{-1} / (\alpha n)  )   ) \, .
\end{equation}
Algebra shows that the optimization problem \eqref{eqn:GMF-approx-limit} has a simple (and unique) closed-form solution when $V_{\theta^*}$ is positive definite; namely:
\begin{equation} \label{eqn:mean-field}
    \mu^*_n = \widehat{\theta}_{\textrm{ML\,-\,$\mathcal{F}_n$}}, \quad \Sigma^*_n = \textrm{diag}(V_{\theta^*})^{-1} / (\alpha n) \,.
\end{equation}
In words, $q(\cdot \, | \, \mu^*_n, \Sigma^*_n)$ is the distribution in the Gaussian mean-field family having the same mean as the limiting distribution of $\pi_{n,\alpha}(\cdot \, | \, X^n)$ but, as we will show, underestimates the covariance. 
We would like to show that the total variation distance between the distributions $\widetilde{\pi}_{n,\alpha}(\cdot\,|\, X^n)$ and $q(\cdot \,|\, \mu^*_n, \Sigma^*_n)$ converges in probability to zero, provided the prior and the likelihood satisfy some regularity conditions. To do this, let $\theta^*$ and $R_{n}(h)$ be defined as in Assumption \ref{asn:A1}.

\begin{assumption}\label{asn:A2}
For any sequence $(\mu_n, \Sigma_n)$ such that $(\sqrt{n}(\mu_n-\theta^*), n \Sigma_n)$ is bounded in $f_{0,n}$-probability, the residual $R_n(h)$ in Assumption \ref{asn:A1} and the prior $\pi$ are such that 
\begin{equation}
\label{eq:asn2.1}
    \int \phi(h \, | \, \sqrt{n}(\mu_n-\theta^*) , n \Sigma_n ) \, \log \left( \frac{\pi(\theta^* + h/\sqrt{n})}{\pi(\theta^*)} \right)dh \rightarrow 0 \, ,
\end{equation}
and
\begin{equation}
\label{eq:asn2.2}
    \int \phi(h \, | \, \sqrt{n}(\mu_n-\theta^*) , n \Sigma_n ) \, R_n(h) \, dh \rightarrow 0 \, .
\end{equation} 
In both cases the convergence is in $f_{0,n}$-probability.
 \end{assumption}

\begin{theorem} 
\label{thm:BVMT-alpha-var}
Suppose that $(\sqrt{n}(\widetilde{\mu}_n-\theta^*), n \widetilde{\Sigma}_n)$ is bounded in $f_{0,n}$-probability where  $(\widetilde{\mu}_n,\widetilde{\Sigma}_n)$ is the sequence defining  $\widetilde{\pi}_{n,\alpha}(\cdot \, | \, X^n)= q(\cdot \, | \, \widetilde{\mu}_n,\widetilde{\Sigma}_n)$. If Assumptions 1 and 2 hold, then
\begin{equation} \label{eqn:BVMT-alpha-var}
     d_{\textrm{TV}}\left( \widetilde{\pi}_{n,\alpha}(\cdot\,|\, X^n), \, q(\cdot \, | \, \mu^*_n, \Sigma^*_n)   \right) \rightarrow 0 \,,
\end{equation}
in $f_{0,n}$-probability, where $\mu_n^*$ and $\Sigma_n^*$ are defined in \eqref{eqn:mean-field}.
\end{theorem}

In words, Theorem 2 shows that the Gaussian mean-field approximation to the $\alpha$-posterior  
 converges to the Gaussian mean-field approximation of asymptotic distribution of the $\alpha$-posterior. Indeed,  the mean and variance parameter of this normal distribution are obtained by \emph{projecting} the limiting distribution obtained in Theorem 1 onto the Gaussian mean-field family. 
 
A detailed proof of Theorem \ref{thm:BVMT-alpha-var} can be found in Appendix \ref{proof:thm2}. A similar result was obtained by \cite{wang2019variational} for the case of $\alpha=1$. Thus, Theorem \ref{thm:BVMT-alpha-var} can be viewed as a generalization of their variational BvM Theorem, applicable to the variational approximations of $\alpha$-posteriors. We note however that our proof technique is quite different from theirs. Indeed, we require a simpler set of assumptions because we restrict ourselves to Gaussian mean-field variational approximations to the $\alpha$-posterior. This enables us to work out a simplified argument that explicitly leverages formulas obtained by computing the  KL divergence between two Gaussians. The key intermediate step in our proof is an asymptotic representation result stated formally in Lemma \ref{lemma3} in Appendix~\ref{section:technical} showing that, under Assumption \ref{asn:A1} and \ref{asn:A2}, for any sequence $(\mu_n, \Sigma_n)$ such that $(\sqrt{n}(\mu_n-\theta^*), n \Sigma_n)$ is bounded in $f_{0,n}$-probability, we have that 
\begin{equation*}
\begin{split}
    &\mathcal{K}( q(\cdot \, | \,\mu_n, \, \Sigma_n  )   \, ||\,  \pi_{n,\alpha}(\cdot  \,| \, X^n) )= \mathcal{K} \left ( q(\cdot \, | \,\mu_n, \, \Sigma_n  )    \, ||\, \phi(\cdot \, | \, \widehat{\theta}_{\textrm{ML\,-\,$\mathcal{F}_n$}}, \, V_{\theta^*}^{-1} / (\alpha n)  ) \right) + o_{f_{0,n}}(1) \,. 
    \end{split}
\end{equation*} 
We use this lemma to show that projecting the $\alpha$-posterior onto the space of Gaussian mean-field distributions is approximately equal to projecting the $\alpha$-posterior's total variation limit in Theorem 1. In particular, the proof of Theorem~\ref{thm:BVMT-alpha-var} shows that
\begin{equation}
    \mathcal{K} ( \widetilde \pi_{n,\alpha}(\cdot \,|\, X^n; \widetilde{\mu}_n, \widetilde{\Sigma}_n )  \, || \, q(\cdot \, | \, \mu^*_n, \Sigma^*_n) ) \to 0 \, ,
\end{equation} 
in $f_{0,n}$-probability. The statement in  \eqref{eqn:BVMT-alpha-var} then follows from the above limit by Pinsker's inequality i.e.\ $d_{TV}(P,Q)\leq \sqrt{2\mathcal{K}(P \: || \:Q)}$ for any two probability distributions $P$ and $Q$. (See part iii) of Lemma B.1 in  \cite{ghosal2017} for a textbook reference on Pinsker's inequality.)

\section{Misspecification Robustness Analysis}
\label{section:Robustness}

This section introduces the measure of robustness we study in this work and our main results related to it. As we will explain below, the main idea is to measure closeness---in terms of KL divergence---of the $\alpha$-posterior or its variational approximations to the correct posterior.   

To formalize this discussion, we introduce a bit more of notation.  In order to analyze model misspecification, following \cite{Gustafson2001}, we  posit a correctly specified parametric model that takes the form $\mathcal{G}_n \equiv \{ g_n( \cdot \, |\, \theta, \gamma): \theta \in \Theta, \gamma \in \Gamma \}$ (notice that under model misspecification, $\mathcal{G}_n$ will differ from $\mathcal{F}_n$).  Correct specification of $\mathcal{G}_n$ here simply means that, for any sample size $n$, there exist parameters $(\theta_n^*,\gamma_n^*) \in \Theta \times \Gamma$ for which $g_n(\cdot \, | \, \theta_n^*, \gamma_n^*)$ equals $f_{0,n}$. Note that the parameter $\theta$ is well-defined in both $\mathcal{F}_n$ and $\mathcal{G}_n$. 

Let $\pi^*$ be a prior over $\Theta \times \Gamma$. We let  $\pi^*_n(\theta \, | \, X^n)$ denote the posterior for $\theta$ based on $\mathcal{G}_n$ and $\pi^*$. When $\mathcal{F}_n$ is misspecified, we refer to $\pi^*_n(\theta \,|\, X^n)$ as the \emph{true} posterior, and when $\mathcal{F}_n$ is correctly specified, the true posterior is simply given by $\pi_{n,\alpha}$ evaluated at $\alpha=1$. In a slight abuse of notation, we denote the ML estimator of $\theta$ based on $\mathcal{G}_n$ simply as $\widehat{\theta}_{\textrm{ML}}$.

Our goal is to compute the KL divergence between $\alpha$-posteriors (or their variational approximations) and the correct posterior. To do so, we need to consider two cases: one in which $\mathcal{F}_n$ is misspecified (in the sense that $f_{0,n} \notin \mathcal{F}_n$) and another in which $\mathcal{F}_n$ is correctly specified. We assume that the decision maker does not know ex-ante whether the model is correctly specified or not, and we denote by $\epsilon_n$ the probability of being misspecified (consequently, $1-\epsilon_n$ denotes the probability of correct specification).

To assess the robustness of $\alpha$-posteriors our interest is to compute the expected value of KL divergence:
\begin{equation} 
\begin{split}\label{eqn:Ratio}
r_n(\alpha) \equiv 
     \epsilon_n \mathcal{K} ( \pi^*_n(\theta | X^n) \: || \: \pi_{n,\alpha}(\theta | X^n) ) + (1-\epsilon_n) \mathcal{K} ( \pi_{n,1}(\theta | X^n) \: || \: \pi_{n,\alpha}(\theta | X^n)  ) \, .
     \end{split}
\end{equation}
The first KL divergence term on the right side of \eqref{eqn:Ratio} measures how difficult it is to distinguish between the \emph{true} posterior under model misspecification and the $\alpha$-posterior. The second is the KL divergence between the regular posterior ($\alpha=1$) and the $\alpha$-posterior. These terms are weighted by the probability of misspecification. Values of $\alpha$ that lead to smaller values of $r_{n}(\alpha)$ are said to be more robust to parametric specification, as the corresponding $\alpha$-posterior is `closer' to the true posterior. To the best of our knowledge, the idea of using the KL divergence between reported posteriors and true posteriors was first introduced by  \cite{Gustafson2001}.

Likewise, we could analyze the robustness of  variational $\alpha$-posteriors by studying:
\begin{equation}
\begin{split}
\label{eqn:Ratio_variational}
  \widetilde{r}_n(\alpha) \equiv  \epsilon_n \mathcal{K} ( \pi^*_n(\theta | X^n) \: || \: \widetilde{\pi}_{n,\alpha}(\theta | X^n)  ) + (1-\epsilon_n) \mathcal{K} ( \pi_{n,1}(\theta | X^n) \: || \: \widetilde{\pi}_{n,\alpha}(\theta | X^n)  ) \, ,
        \end{split}
\end{equation}
{where \eqref{eqn:Ratio_variational} is the same as \eqref{eqn:Ratio}, except we have replaced the $\alpha$-posterior $\pi_{n,\alpha}(\theta | X^n)$ in \eqref{eqn:Ratio} with its variational approximation $\widetilde{\pi}_{n,\alpha}(\theta | X^n)$ in \eqref{eqn:Ratio_variational}.}
Both $r_n(\alpha)$ and $\widetilde{r}_n(\alpha)$ are  random variables (where the randomness comes from the sampled data $X^n$ used to construct the posterior distributions) and their magnitudes depend on the structure of the correctly and incorrectly specified models, on the priors, and on the sample size. We are able to make progress on the analysis of \eqref{eqn:Ratio} and \eqref{eqn:Ratio_variational} by relying on asymptotic approximations to the infeasible posterior, the regular posterior, and the $\alpha$-posterior  and its variational approximation.

It is well known that the Bernstein-von Mises theorem for correctly specified models---e.g., \cite{dasgupta2008asymptotic}, p.\ 291---implies that under some regularity conditions on the statistical model $\mathcal{G}_n$ and the prior $\pi^*$ (analogous to Assumption \ref{asn:A1} and \eqref{eqn:concentration}), the true, infeasible posterior $\pi^*_n(\theta | X^n)$ is close in total variation distance to the p.d.f.\ of a $\mathcal{N} ( \widehat{\theta}_{\textrm{ML}}, \Omega / n )$ random variable. Theorems \ref{thm:BVMT-alpha} and \ref{thm:BVMT-alpha-var}, naturally suggest surrogates for \eqref{eqn:Ratio} and \eqref{eqn:Ratio_variational} where we replace the densities by their asymptotically normal approximations. We denote the surrogate measures by $r_n^*(\alpha)$ and $\widetilde{r}_n^*(\alpha)$. Define
\begin{equation}\label{eqn:alpha_robust}
    \alpha_n^* \equiv \arg \min_{\alpha \geq 0} \, r_n^*(\alpha) \, , \qquad \widetilde{\alpha}_n^* \equiv \arg \min_{\alpha \geq 0} \,  \widetilde{r}_n^*(\alpha) \, ,
\end{equation}
where 
\begin{equation}  
\begin{split}
\label{eq:rn_star}
    r_n^*(\alpha) & \equiv  \epsilon_n \mathcal{K} \left( \phi(\cdot\,|\, \widehat{\theta}_{\textrm{ML}},\Omega /n) \: || \: \phi(\cdot\,|\, \widehat{\theta}_{\textrm{ML\,-\,$\mathcal{F}_n$}}, V^{-1}_{\theta^*}/(\alpha n)) \right) \\ 
    &\quad +  (1-\epsilon_n) \mathcal{K} \left( \phi(\cdot\,|\, \widehat{\theta}_{\textrm{ML\,-\,$\mathcal{F}_n$}}, V^{-1}_{\theta^*}/n)  \: || \: \phi(\cdot\,|\, \widehat{\theta}_{\textrm{ML\,-\,$\mathcal{F}_n$}}, V^{-1}_{\theta^*}/(\alpha n)) \right) \, ,
\end{split}
\end{equation}
and
\begin{equation}  
\begin{split}
\label{eq:tilde_rn_star}
    \widetilde{r}_n^*(\alpha) & \equiv \epsilon_n \mathcal{K} \left( \phi(\cdot\,|\, \widehat{\theta}_{\textrm{ML}},\Omega /n) \: || \: \phi(\cdot\,|\, \widehat{\theta}_{\textrm{ML\,-\,$\mathcal{F}_n$}}, \diag({V}_{\theta^*})^{-1}/(\alpha n)) \right) \\
    & \quad + (1-\epsilon_n) \mathcal{K} \left( \phi(\cdot\,|\, \widehat{\theta}_{\textrm{ML\,-\,$\mathcal{F}_n$}}, V^{-1}_{\theta^*}/n)  \: || \: \phi(\cdot\,|\, \widehat{\theta}_{\textrm{ML\,-\,$\mathcal{F}_n$}}, \diag({V}_{\theta^*})^{-1}/(\alpha n)) \right) \, .
\end{split}
\end{equation}

\begin{theorem} \label{thm:limit_r}
Let $p \equiv dim(\theta)$, let $V_{\theta^*}$ be the positive definite matrix satisfying Assumption \ref{asn:A1}, and let $\widetilde{V}_{\theta^*} \equiv \textrm{diag}(V_{\theta^*})$.  
Suppose $\widehat{\theta}_{\textrm{ML}}\to \theta_0$ in $f_{0,n}$-probability. If $\theta_0 \neq \theta^*$ and $n \epsilon_n \rightarrow \varepsilon \in (0,\infty)$ then
\begin{eqnarray}
\alpha^*_n  &\to& \frac{p}{p + \varepsilon (\theta_0-\theta^*)^\top V_{\theta^*}(\theta_0-\theta^*)   }<1 \, ,  \label{eqn:T3-1} \\
\widetilde{\alpha}^*_n  &\to& \frac{p}{\tr(  \widetilde{V}_{\theta^*} V^{-1}_{\theta^*}) +  \varepsilon (\theta_0-\theta^*)^\top \widetilde{V}_{\theta^*}(\theta_0-\theta^*) }<1 \, ,
\label{eqn:T3-2}
\end{eqnarray}
where the convergence is in $f_{0,n}$-probability.
\end{theorem}

We now discuss the meaning and implications of Theorem 3. 
Equation \eqref{eqn:T3-1} says that, for a properly tuned value of $\alpha$, the $\alpha$-posterior is---with high probability and in large samples---more robust than the regular posterior. The limit of $\alpha^*_n$ suggests that the properly tuned value of $\alpha$ decreases as the probability of misspecification increases. This makes conceptual sense: it seems reasonable to down-weight the likelihood if it is known that, very likely, it is misspecified. On the other extreme, if the likelihood is known to be correct there is no gain from down-weighting the likelihood, as this simply creates a difference in the scaling of the $\alpha$-posterior relative to the true posterior. 

The formula also says that if the misspecification implied by the model is large (the difference between $\theta_0$ and $\theta^*$ is large), then the properly tuned value of $\alpha$ must be small. 

Equation \eqref{eqn:T3-2} refers to the properly calibrated value of $\alpha$ for the variational approximations of $\alpha$-posteriors. The analysis is similar to what we have already discussed, but here we need to take into account that variational approximations tend to further distort the variance matrix. Indeed, Theorem \ref{thm:BVMT-alpha-var} has shown that the asymptotic variance of the variational approximations is $\widetilde{V}_{\theta^*}^{-1}/\alpha n$, as opposed to $V_{\theta^*}^{-1} / \alpha n$, where  we have defined $\widetilde{V}_{\theta^*} \equiv \textrm{diag}(V_{\theta^*})$, for positive definite $V_{\theta^*}$. %
It is well known that the variational approximation understates the variances of each coordinate of $\theta$ (\cite{blei2017variational}), hence
$[( \widetilde{V}_{\theta^*})^{-1}]_{jj} \leq [V_{\theta^*}^{-1}]_{jj}$ for $j=1, \ldots, p$. Therefore,
$\textrm{tr}(  \widetilde{V}_{\theta^*} V^{-1}_{\theta^*}) \geq p,$
which establishes the inequality in \eqref{eqn:T3-2}.

There is one additional derivation associated to the formulae in Theorem 3. Consider the optimized expected value of the KL divergence based on the asymptotic approximations to $\alpha$-posteriors and their variational approximations. From the definition of $r_n^*$ and Theorem \ref{thm:limit_r}:
\begin{equation}
    {2\lim_{n\to\infty}r_n^*(\alpha_n^*)=} -p \log (p) + p \log \left(p + \varepsilon (\theta_0 - \theta^*)^\top V_{\theta^*} (\theta_0 - \theta^*) \right) \, . 
\end{equation}
Likewise,
\begin{equation} \label{eqn:optimized-var}
2\lim_{n\to\infty}\widetilde{r}_n^*(\widetilde{\alpha}_n^*)=     -p \log (p) + p \log \left( \textrm{tr}\left(  \widetilde{V}_{\theta^*} V^{-1}_{\theta^*} \right) + \varepsilon (\theta_0 - \theta^*)^\top \widetilde{V}_{\theta^*} (\theta_0 - \theta^*) \right) + \log \left (\frac{|V_{\theta^*}|}{|\widetilde{V}_{\theta^*}|} \right) \,. 
\end{equation}
Compare these two equations with the expected KL of the usual posterior $(\alpha=1)$. In this case, the KL under correct specification equals zero, so that the only relevant piece is the KL distance under misspecification. Algebra shows that
\[ 2\lim_{n \rightarrow \infty} r_n^*(1) = \varepsilon (\theta_0 - \theta^*)^\top V_{\theta^*} (\theta_0 - \theta^*) \, .   \]
Thus, the expected KL distance for the regular posterior increases linearly in the magnitude of the misspecification (the distance between $\theta_0$ and $\theta^*$). The optimized KL for both the $\alpha$-posteriors and their variational approximations is also monotonically increasing in this term, but its growth is logarithmic.   

One final remark concerns the well-known result that the asymptotic variance of Bayesian posteriors under misspecified models does not coincide with the asymptotic `sandwich' covariance matrix of the Maximum Likelihood Estimator under misspecification. This suggests that instead of targeting the true (perhaps infeasible) posterior, it might make more sense to target the artificial posterior suggested by \cite{muller2013risk}, which is a normal centered at the Maximum Likelihood estimator but with sandwich covariance matrix. This choice of target distribution does not change our results. The reason is that under our assumptions the part of the expected Kullback Leibler under misspecification only depends on the difference between true and pseudo-true parameter and the covariance matrix of the reported posterior.

\section{Illustrative Example}
\label{section:Example}
To illustrate our main results, this section studies an example of model misspecification in the form of a linear regression model with omitted variables. Our objective is twofold. First, we present a simple environment where the high-level assumptions of our main theorems can be easily verified and discussed. Second, an appropriate choice of priors in this model yields closed-form solutions for the $\alpha$-posteriors and their (Gaussian mean-field) variational approximations. Thus, it is possible to provide additional details about the nature of the Bernstein-von Mises theorems that we have established, as well as the optimal choice of $\alpha$.

\subsection{True and misspecified model}

Consider a random sample of an outcome variable $Y_i$ with control variables $W_i \in \mathbb{R}^{p}$ and $Z_i \in \mathbb{R}^{d}$. The true data generating process is an homoskedastic Gaussian linear regression model
\[ Y_i = \theta_0^\top W_i + \gamma_0^\top Z_i + \varepsilon_i \,,  \]
where $\epsilon_i \sim \mathcal{N}(0, \sigma^2_{\epsilon})$ independently of $W_i$ and $Z_i$. The joint distribution of $W_i$ and $Z_i$ is assumed to have a density $h(w_i,z_i)$ with respect to the Lebesgue measure on $\mathbb{R}^{p+d}$.  

The statistician faces an omitted variables problem, in that she would like to estimate $\theta_0$ but only observes $(Y_i, W_i)$. The statistician's misspecified model posits that
\[ Y_i = \theta^\top W_i + u_i \, ,  \]
where $u_i$ is assumed to be univariate normal with mean zero and a pressumedly known variance $\sigma^2_{u}$ independently of $W_i$. For simplicity, we assume that the statistician has correctly specified the marginal distribution of $W_i$.  

\subsection{Pseudo-true parameter and LAN assumption}

If we denote the data as $X^n \equiv \{ (Y_i, W_i)\}_{i=1}^{n}$, the likelihood is 
\[  f(X^n \, | \, \theta) = \frac{1}{(2\pi \sigma^2_u)^{n/2}} \exp \left( -\frac{1}{2 \sigma^2_u } \sum_{i=1}^{n} (Y_i - \theta^\top W_i)^2 \right) \prod_{i=1}^{n} h(W_i) \, ,   \]
and the Maximum Likelihood estimator is simply the least-squares estimator of $\theta$:
\[ \widehat{\theta}_{ML} = \left( \frac{1}{n} \sum_{i=1}^{n} W_i W_i^\top \right)^{-1} \frac{1}{n} \sum_{i=1}^{n} W_i Y_i \, . \]
It is straightforward to show that under mild assumptions on the joint distribution of $(W_i,Z_i)$, the Maximum Likelihood estimator is $\sqrt{n}$-asymptotically normal around the pseudo-true parameter:
\[ \theta^* \, \equiv  \, \theta_0 + \left( \mathbb{E}[W_i \, W_i^\top]^{-1} \,  \mathbb{E}[W_i \, Z_i^\top] \right) \gamma_0 \, , \]
which equals the true parameter, $\theta_0$, plus the usual omitted variable bias formula. Algebra shows that as long as the sample second moments of $W_i$ converge in probability (under the true model) to the positive definite matrix $\mathbb{E}[W_i \, W_i^\top]$, the stochastic LAN assumption is satisfied with 
\[ V_{\theta^*} \, \equiv \, \mathbb{E}[W_iW_i^\top]/\sigma_u^2 \,. \]
This means that, in this example, the curvature of the likelihood around the pseudo-true parameter does not depend on $\theta^*$. 

\subsection{$\alpha$-posteriors and their variational approximations}
Consider now the setting where we assume a commonly used Gaussian prior, denoted $\pi$, for $\theta$. In particular, suppose
\[ \theta \sim \mathcal{N}(\mu_{\pi}, \, \sigma^2_u \, \Sigma^{-1}_{\pi}) \,. \]
The computation of the $\alpha$-posterior in this set-up is straightforward, as the $\alpha$-power of the Gaussian likelihood is itself Gaussian with the new scale divided by $\alpha$. Thus, algebra shows that the $\alpha$-posterior for the linear regression model is also multivariate normal with mean parameter 
\begin{equation} \label{eqn:alpha-mean-linref}
    \mu_{n,\alpha} \equiv \left( \frac{1}{n} \sum_{i=1}^{n} W_i W_i^\top + \frac{1}{\alpha n } \, \Sigma_{\pi} \right)^{-1} \left( \frac{1}{\alpha n } \, \Sigma_{\pi} \mu_\pi + \frac{1}{n} \sum_{i=1}^{n} W_i Y_i\right) \, , 
\end{equation}
and covariance matrix
\begin{equation}  \label{eqn:alpha-variance-linref}
  \Sigma_{n,\alpha} \equiv  \frac{\sigma^2_u}{ \alpha n }\left( \frac{1}{n} \sum_{i=1}^{n} W_i W_i^\top + \frac{1}{\alpha n } \,  \Sigma_{\pi} \right)^{-1} \, .
\end{equation}
Because we have closed-form solutions for the $\alpha$-posteriors, one can readily show that they concentrate at rate $\sqrt{n}$ around $\theta^*$ for any fixed $(\alpha,\mu_{\pi},\Sigma_{\pi})$. This is shown in Appendix \ref{section:B1}.   

Since the assumptions of Theorem \ref{thm:BVMT-alpha} are met, then the total variation distance between the $\alpha$-posterior and the multivariate normal
\begin{equation} \label{eqn:limit_regression}
\mathcal{N} \left( \widehat{\theta}_{ML}, \, \frac{\sigma^2_{u}}{\alpha n} \, \mathbb{E}[W_i \, W_i^\top]^{-1} \right) \,,
\end{equation}
must converge in probability to zero. In fact, in this example, it is possible to establish a stronger result: the KL distance between $\pi_{n,\alpha}$ and the distribution in $\eqref{eqn:limit_regression}$ converges in probability to zero (see Appendix \ref{section:B2}). This example thus raises the question of whether \emph{entropic} Bernstein-von Mises theorems are more generally available for $\alpha$-posteriors in misspecified models.\footnote{\cite{clarke1999asymptotic} showed that---in a smooth parametric model with a well-behaved prior---the relative entropy between a posterior density and an appropriate normal tends to zero in probability and in mean. If an analogous result were available for standard posterior distributions in misspecified parametric models, it might be possible to extend it to cover $\alpha$-posteriors. }

This simple linear regression example also shows that the Bernstein-von Mises theorem is not likely to hold if $\alpha$ is chosen in a way that approaches zero very quickly. 
In particular, consider a sequence $\alpha_n$ for which $\alpha_n n$ converges to a strictly positive constant. Then, in this simple example the total variation distance between the $\alpha_n$-posterior and the distribution in \eqref{eqn:limit_regression} will be  bounded away from zero. This is shown in Appendix \ref{section:B3}.

Finally, in this example the mean-field Gaussian variational approximation of the $\alpha$-posterior also has a closed-form expression. Algebra shows that the variational approximation has exactly the same mean as the $\alpha$-posterior, but variance equal to 
\begin{equation}
    \widetilde{\Sigma}_{n,\alpha} \equiv  \frac{\sigma^2_u}{ \alpha n } \left( \textrm{diag} \left( \frac{1}{n} \sum_{i=1}^{n} W_i W_i^\top + \frac{1}{\alpha n } \, \Sigma_{\pi} \right) \right)^{-1} \, .
\end{equation}
In this example, it is possible to show that the Bernstein-von Mises theorem holds for the variational approximation to the $\alpha$-posterior (not only in total variation distance, but also in KL divergence). The assumptions of Theorem \ref{thm:BVMT-alpha-var} are verified in Appendix \ref{section:B4}.

\subsection{Expected KL and Optimal $\alpha$} 
Finally, we discuss the expected KL criterion and the  choice of $\alpha$ in the context of our example. In Section \ref{section:Robustness}, we defined $r_n(\alpha)$ as the expected KL divergence between the true posterior of $\theta$ and the $\alpha$-posterior. As we therein explained, this expected KL measure is typically difficult to compute because---with the exception of some stylized examples such as our linear regression model with omitted variables---the $\alpha$-posteriors and their variational approximations are not available in closed-form. 

For this reason, we chose to work instead with a surrogate measure $r_n^*(\alpha)$, which, motivated by the Bernstein-von Mises theorem, replaces the true posterior and the $\alpha$-posterior by their asymptotic approximations. In our linear regression example, it is possible to formalize the relationship between $r_{n}(\alpha)$ and $r^*_{n}(\alpha)$. Algebra shows that for any fixed $\alpha$ and any Gaussian prior, $r_n(\alpha)-r_n^*(\alpha) \to 0$ as $n \to \infty,$
as one would have expected (see Appendix \ref{section:B6}). One of the challenges in generalizing this result is that the Bernstein-von Mises theorems we have established are in total variation, thus we cannot use them directly to analyze the behavior of the expected KL divergence. 

Regarding the choice of $\alpha$, we have shown in Theorem \ref{thm:limit_r} that the limit of the optimal choice is
$$ \alpha^* = \frac{p}{p + \varepsilon (\theta_0-\theta^*)^\top V_{\theta^*}(\theta_0-\theta^*)   } \, , $$
under the assumption that the true parameter, $\theta_0$, is different from the pseudo-true parameter $\theta^*$. In our example, this happens whenever $\gamma_0 \neq 0$ (i.e., there are indeed omitted variables) and the omitted variables are correlated with the observed controls (i.e., when the omitted variable bias is different from zero). 

Since in the linear regression example it is possible to compute $r_{n}(\alpha)$ explicitly, it is also possible to choose $\alpha$ to minimize this expression. Algebra shows that for any $\alpha' \neq \alpha^*$, we have that  
$ r_n(\alpha^*) < r_n(\alpha')$
for sufficiently large $n$.

\section{Concluding Remarks and Discussion}
\label{section:Conclusion}

In this work, we have studied the robustness to model misspecification of $\alpha$-posteriors and their variational approximations with a focus on parametric, low-dimensional models.  
To formalize the notion of robustness we built on the seminal work of \cite{Gustafson2001} and his suggested measure of sensitivity to parametric model misspecification. To state it simply, if two different procedures both lead to  incorrect a posteriori inference (either due to model  misspecification or computational considerations), one procedure is more robust (or less sensitive) than  the other if it is  closer---in terms of KL divergence---to the \emph{true} posterior. Thus, we analyzed the KL divergence between true posteriors and the distributions reported by either the $\alpha$-posterior approach or their variational approximations. 

Obtaining general results about the properties of the KL divergence between true and reported posteriors is quite challenging, as this will typically depend on the priors, the data, the statistical model, and the form of misspecification. We were able to make progress by relying on asymptotic approximations to $\alpha$-posteriors and their variational approximations. 

In particular, we established a  Bernstein-von Mises (BvM) theorem in total variation distance for $\alpha$-posteriors (Theorem \ref{thm:BVMT-alpha}) and for their (Gaussian mean-field) variational approximations (Theorem \ref{thm:BVMT-alpha-var}). Our results provided a generalization of the results in  \cite{wang2019variational,WangBleiJASA2019}, who focus on the case in which $\alpha = 1$. We also extend the results of \cite{Yu2019}, who establish the BvM theorem for $\alpha$-posteriors under a weaker norm (weak convergence), but under more primitive conditions. 

We think these asymptotic approximations have value per se. For example, we learned that relative to the BvM theorem for the standard posterior or its variational approximation, the choice of $\alpha$ only re-scales the limiting variance. The new scaling acts as if the observed sample size were $\alpha \cdot n$ instead of $n$, but the location for the Gaussian approximation continues to be the Maximum Likelihood estimator. Since choosing $\alpha<1$ inflates the $\alpha$-posterior's variance relative to the usual posterior, then the tempering parameter corrects some of the variance understatement of standard variational approximations to the posterior. Also, there is some recent work considering variational approximations using the $\alpha$-R\'enyi divergence instead of the KL divergence \citep{rao2020}. It might be interesting to explore whether it is possible to derive a result analogous to Theorem \ref{thm:BVMT-alpha-var}, where we approximate the limiting distribution of $\alpha$-R\'enyi approximate posteriors by projecting the limiting distribution obtained in Theorem \ref{thm:BVMT-alpha}.

The main use of the asymptotic approximations in our paper, however, was simply to facilitate the computation of the suggested measure of robustness. This required elementary calculations once we have multivariate Gaussians with parameters that depend on the data, the sample size, and the `curvature' of the likelihood. 

An important caveat of our results is that we focused on analyzing \emph{the KL divergence between the limiting distributions, as opposed to the limit of KL divergence between reported and true posteriors}. Although in some simple models the two are equivalent (for example, a linear regression model with omitted variables and Gaussian priors), the general result requires further exploration. Unfortunately, we do not yet have a good solution. It is easy to show that the function $(P,Q) \mapsto \mathcal{K}(P||Q)$ is lower semi-continuous in total variation distance, so it might be possible to get a bound on the KL divergence we want to study with the KL divergence of the Gaussian limits. However, the interesting part is not the divergences themselves, but rather the $\alpha$'s that minimize them. We think that perhaps the use of the Theorem of the Maximum \cite{berge1963topological} and its generalizations could be useful for this analysis. It is possible that the continuity results can be strengthened in our case, since we are dealing with Gaussians in the limit,
but we do not yet have answers. Also, a formal analysis might require the derivation of \emph{entropic} BvM theorems, where the distance is measured using KL divergence, as in \cite{clarke1999asymptotic}. 

Finally, even though our paper has a theoretical prescription for choosing the tempering parameter $\alpha$, further research is needed to translate this into a practical recommendation. As discussed in detail, our calculations suggest that $\alpha^*_n$ tends to be smaller as both the probability of misspecification $\epsilon_n$ and the difference between the true and pseudo-true parameters increase. It might be possible to hypothesize some value for the probability of misspecification. However, the true parameter is not known (and cannot be estimated consistently under misspecification). We leave the question of how to optimally choose the tempering parameter for $\alpha$-posterior and their approximations for future research. One promising line of work is to do a full Bayesian treatment of the problem as in \cite{pmlr-v70-wang17g}.

\acks{We would like to thank Pierre Alquier, St\'{e}phane Bonhomme, Yun Ju, Debdeep Pati, and Yixin Wang for extremely helpful comments and suggestions. All errors remain our own.
Cynthia Rush would like to acknowledge support for this project
from the National Science Foundation (NSF CCF \#1849883), the Simons Institute for the Theory of Computing, and NTT Research. }

%\newpage
\appendix
\section{Proofs of Main Results}
\label{section:Main_Appendix}

%%%%%%%%%%%%%%  Proof of Theorem 1 %%%%%%%%%%%%%%%%%%%
\subsection{Proof of Theorem 1}\label{proof:thm1}

This proof follows Theorem 2.1 in \cite{kleijn2012bernstein}, which shows that the posterior under misspecification is asymptotically normal, but our proof is adapted and simplified for the $\alpha$-posterior framework. In what follows, we let $\Delta_{n,\theta^*} \equiv \sqrt{n}\,(\widehat{\theta}_{\textrm{ML\,-\,$\mathcal{F}_n$}}-\theta^*)$ as in Assumption \ref{asn:A1}.

By the Theorem 1 assumption that $\theta^*$ is in the interior of $\Theta \subset \mathbb{R}^p$, there exists a sufficiently small $\delta>0$ such that the open ball $B_{\theta^*}(\delta) \equiv \{ \theta: || \theta-\theta^*|| < \delta\}$ is a neighborhood of the (pseudo-)true parameter $\theta^*$ in  $\Theta$. In particular, we choose $\delta$ such that $B_{\theta^*}(\delta)$ belongs to the neighborhood of $\theta^*$ in which it is assumed $\pi$ is continuous and positive. 
Note that for any compact set $K_0 \subset \mathbb{R}^p$ including the origin, we can find an integer $N_0 \equiv N_0(K_0,B_{\theta^*}(\delta))$ sufficiently large such that for any vector $h \in K_0$, we have that the perturbation of $\theta^*$ in the direction $h/\sqrt{n}$, meaning $\theta^* + h/\sqrt{n}$, belongs to $B_{\theta^*}(\delta)$ whenever $n \ge N_0$.

The goal is to show that the total variation distance between the $\alpha$-posterior of $\theta$, denoted   $\pi_{n,\alpha}( \cdot \, | \, X^n )$, and a multivariate Normal distribution with mean $\widehat{\theta}_{\textrm{ML}-\mathcal{F}_n}$ and variance $V_{\theta^*}^{-1}/(n \alpha)$ goes to zero in $f_{0,n}$-probability. Because the total variation distance is invariant to the simultaneous re-centering and scaling of both measures being compared (\cite{van2000asymptotic}), it is more convenient to work with the $\alpha$-posterior of the transformation $\sqrt{n} \, (\theta-\theta^*)$ and compare it to the similarly re-centered and scaled multivariate Normal distribution, i.e.\ one with mean $\Delta_{n,\theta^*}$ and variance $V_{\theta^*}^{-1}/\alpha$.

For vectors $g, h \in K_0$, the following random variable will be used to bound the average total variation distance between the $\alpha$-posterior and the alleged multivariate normal limit:
\begin{equation}
\label{eq:fn_theorem_def}
 f_n(g,h) \equiv \Big \{1 - \frac{\phi_n(h) }{\pi_{n,\alpha}^{LAN}(h\:| \:X^n)} \, \frac{\pi_{n,\alpha}^{LAN}(g\:| \:X^n)}{\phi_n(g)} \Big \}^+ \, ,
\end{equation}
where $\phi_n(h) \equiv n^{-1/2}\phi(h\,|\, \Delta_{n,\theta^*},\: V_{\theta^*}^{-1}/\alpha)$ and $\pi_{n,\alpha}^{LAN}(h\:| \: X^n) \equiv n^{-1/2} \pi_{n,\alpha}(\theta^*+h/\sqrt{n}\:| \: X^n)$, are scaled versions of the densities that we want to compare using the total variation distance, and $\{x\}^+ = \max\{0, x\}$ denotes the positive part of $x$. Define also $\pi_n(h) \equiv n^{-1/2}\, \pi(\theta^* + h/\sqrt{n})$ to be the density of the prior distribution of the transformation $\sqrt{n}\,(\theta-\theta^*)$. It follows that $f_n$ in \eqref{eq:fn_theorem_def} is well-defined on $K_0 \times K_0$ for all $n>N_0$, as in this regime $\pi_{n,\alpha}^{LAN}(h\:| \:X^n)$ is guaranteed to be positive since $\theta^* + h/\sqrt{n}$ belongs to $B_{\theta^*}(\delta)$ as discussed above.

Let $\overline B_{\bf{0}}(r_n)$ denote a closed ball of radius $r_n$ around $\bf{0}$. Since $d_{TV} \leq 1$ and the expectation is linear, we have that  for any sequence $r_n$ and for any $\eta>0$:
\begin{equation} 
\begin{split}
\label{eqn:aux1_T1}
 &\mathbb{E}_{f_{0,n}} \left [ d_{\textrm{TV}}\left( \pi_{n,\alpha}^{LAN}(\cdot \:| \:X^n), \, \phi_n(\cdot)   \right) \right]  \\
 & \leq \mathbb{E}_{f_{0,n}} \left [d_{\textrm{TV}}( \pi_{n,\alpha}^{LAN}(\cdot\:| \: X^n), \, \phi_n(\cdot)) \mathbf{1} \left\{\sup_{g,h \, \in \, \overline B_{\bf{0}}(r_n)} f_n(g,h) \le \eta \right\} \right] +  \mathbb{P}_{f_{0,n}}  \left ( \sup_{g,h \, \in \, \overline B_{\bf{0}}(r_n)} f_n(g,h) > \eta \right ) \, .
 \end{split}
\end{equation}
The proof is completed by bounding the two terms on the right side of \eqref{eqn:aux1_T1}.

First we bound the expectation on the right side of \eqref{eqn:aux1_T1}. Lemma \ref{lemma1} in Appendix \ref{section:technical} implies 
\begin{equation} 
\begin{split}
d_{\textrm{TV}}& ( \pi_{n,\alpha}^{LAN}(\cdot\:| \: X^n), \, \phi_n ) \le \sup_{g,h \, \in \, \overline B_{\bf{0}}(r_n)} f_n(g,h)  + \int_{||h|| > r_n} \pi_{n,\alpha}^{LAN}(\cdot\:| \: X^n) \, dh + \int_{||h||>r_n} \phi_n(h) \, dh \, ,
\end{split}
\end{equation}
therefore 
\begin{equation}
    \begin{split}
    \label{eq:above_eq}
    &\mathbb{E}_{f_{0,n}} \left [d_{\textrm{TV}}( \pi_{n,\alpha}^{LAN}(\cdot\:| \: X^n), \, \phi_n(\cdot)) \mathbf{1} \left\{\sup_{g,h \, \in \, \overline B_{\bf{0}}(r_n)} f_n(g,h) \le \eta \right\} \right] \\
    &\qquad \leq \eta  + \mathbb{E}_{{f_{0,n}}} \left[ \int_{{||h||} >r_n} \pi_{n,\alpha}^{LAN}(h\:| \:X^n) \, dh \right]  +  \mathbb{E}_{{f_{0,n}}} \left[ \int_{{||h||} >r_n} \phi_n(h)\,  dh \right] \, .
    \end{split}
\end{equation}
The bound in \eqref{eq:above_eq} uses that by the non-negativity of $\phi_n(\cdot)$, which implies
\[ \mathbb{E}_{f_{0,n}} \left [\int_{{||h||} >r_n} \phi_n(h)\,  dh \,\, \mathbf{1} \left\{\sup_{g,h \, \in \, \overline B_{\bf{0}}(r_n)} f_n(g,h) \le \eta \right\} \right] \leq\mathbb{E}_{f_{0,n}} \left [\int_{{||h||} >r_n} \phi_n(h)\,  dh \right] \, ,\]
and a similar upper bound for the third term on the right side of \eqref{eq:above_eq}.
In addition, by the concentration assumption of the theorem (defined in equation  \eqref{eqn:concentration}), there exists an integer $N_1(\eta, \epsilon)$ % > N(\eta, \epsilon)$ 
such that, for all $ n > N_1(\eta, \epsilon)$,
\begin{equation}
\mathbb{E}_{f_{0,n}} \left [ \int_{||h|| >r_n} \pi_{n,\alpha}^{LAN}(h\,|\,X^n) \, dh \right] < \epsilon.
\label{eq:bound2}
\end{equation}
Also, by Lemma 5.2 in \cite{kleijn2012bernstein} which exploits properties of the multivariate normal distribution, there exists an integer $N_2(\eta, \epsilon)$, such that for all $ n > N_2(\eta, \epsilon)$,
\begin{equation}
\mathbb{E}_{f_{0,n}} \left [ \int_{||h || >r_n} \phi_{n}(h) \, dh \right ] < \epsilon \, .
\label{eq:bound3}
\end{equation}
Now plugging \eqref{eq:bound2} and \eqref{eq:bound3} into \eqref{eq:above_eq}, defining $\widetilde{N}(\eta, \epsilon) = \max\{N_1(\eta, \epsilon), N_2(\eta, \epsilon)\}$, we find for all $n > \widetilde{N}(\eta, \epsilon)$,
\begin{equation}
    \begin{split}
    \label{eq:above_eq2}
    \mathbb{E}_{f_{0,n}}& \left [d_{\textrm{TV}}\left( \pi_{n,\alpha}^{LAN}(\cdot\:| \: X^n), \, \phi_n(\cdot)   \right) 1 \left\{\sup_{g,h \, \in \, \overline B_{\bf{0}}(r_n)} f_n(g,h) \le \eta \right\} \right] \leq \eta +  2\epsilon.
    \end{split}
\end{equation}

Lemma \ref{lemma2} in Appendix \ref{section:technical} shows that, for a given $\eta, \epsilon >0$, there exists a sequence $r_n \to +\infty$ and $N(\eta,\epsilon)$ such that the second term on the right side of equation \eqref{eqn:aux1_T1} is small for $n > N(\eta,\epsilon)$; that is, for all $ n > N(\eta, \epsilon)$,
\begin{equation}
\mathbb{P}_{f_{0,n}} \left (  \sup_{g,h \, \in \, \overline B_{\bf{0}}(r_n)} f_n(g,h) > \eta \right ) \le \epsilon \, .
\label{eq:bound1}
\end{equation}
We mention that it is in the proof of Lemma \ref{lemma2} that we make use the stochastic LAN condition in Assumption \ref{asn:A1}.

Finally we conclude from \eqref{eqn:aux1_T1}, using the bounds in \eqref{eq:above_eq2} and  \eqref{eq:bound1}, that for all $n > \max\{N(\eta, \epsilon), \widetilde{N}(\eta, \epsilon)\}$,
\begin{equation*}
\begin{split}
\mathbb{E}_{f_{0,n}} \Big [d_{\textrm{TV}}\left( \pi_{n,\alpha}^{LAN}(\cdot\,|\,X^n), \, \phi_n(\cdot)   \right) \Big] \le \eta +  2\epsilon +  \epsilon = \eta +  3\epsilon \, .
\label{eq:expectation_bound}
\end{split}
\end{equation*}
A standard application of Markov's inequality gives the desired result.

%%%%%%%%%%%%%%  Proof of Theorem 2 %%%%%%%%%%%%%%%%%%%
\subsection{Proof of Theorem 2}\label{proof:thm2}
Let $\widetilde \pi_{n,\alpha}(\cdot \,|\, X^n )  =q(\cdot\,|\, \widetilde{\mu}_n, \widetilde{\Sigma}_n)$ be the \emph{Gaussian mean-field approximation to the $\alpha$-posterior},  defined in \eqref{eqn:GMF-approx}.
We will prove that 
\begin{equation}
    \label{eq:KL_convergenceThm2}
    \mathcal{K} \left( \widetilde \pi_{n,\alpha}\left(\cdot \,|\, X^n \right)  \: || \: q(\cdot \,| \, \mu_n^*, \Sigma_n^*)
     \right) =  \mathcal{K} \left( \phi(\cdot\,|\, \widetilde{\mu}_n, \widetilde{\Sigma}_n)  \: || \: \phi(\cdot \, | \, \widehat{\theta}_{\textrm{ML\,-\,$\mathcal{F}_n$}}, \, \diag(\alpha nV_{\theta^*})^{-1}  )  \right) \to 0 \, ,
\end{equation} 
in $f_{0,n}$-probability where the equality follow since $q(\cdot \,| \, \mu_n^*, \Sigma_n^*)$ is Gaussian with mean and covariance defined in \eqref{eqn:mean-field}.
The statement in  \eqref{eqn:BVMT-alpha-var} follows by Pinsker's inequality  as it ensures that convergence in Kullback-Leibler divergence implies convergence in total variation distance.

Note that the KL divergence between two $p$-dimensional Gaussian distributions can be computed explicitly as
\begin{align}
\label{eq:KLnormals}
    &\mathcal{K}\left( \phi(\cdot \, | \, \mu_1, \,\Sigma_1)\: ||\:  \phi(\cdot \, | \, \mu_2, \,\Sigma_2)\right) =\frac{1}{2} \left[\log\left(\frac{|\Sigma_2|}{|\Sigma_1|}\right) + \tr\left(\Sigma_2^{-1} \Sigma_1\right) + (\mu_2 - \mu_1)^\top \Sigma_2^{-1}(\mu_2 - \mu_1)  - p \right] \, .
\end{align}
Applying the identity \eqref{eq:KLnormals} we see that
\begin{equation*}
    \mathcal{K} \left( \phi(\cdot\,|\, \widetilde{\mu}_n, \widetilde{\Sigma}_n)  \: || \: \phi(\cdot \, | \, \widehat{\theta}_{\textrm{ML\,-\,$\mathcal{F}_n$}}, \, \diag(\alpha nV_{\theta^*})^{-1}  )  \right) =T_1+T_2 \, ,
 \end{equation*}
where
\begin{align}
    \label{eq:T1}
    T_1&=\frac{1}{2}(\widehat{\theta}_{\textrm{ML\,-\,$\mathcal{F}_n$}}-\widetilde \mu_n)^{\top} \text{diag}(\alpha  n V_{\theta^*}) (\widehat{\theta}_{\textrm{ML\,-\,$\mathcal{F}_n$}}-\widetilde \mu_n) \, , \\
    \label{eq:T2}
    T_2 &=
        \frac{1}{2}  \tr( \text{diag}(\alpha  n V_{\theta^*}) \widetilde \Sigma_n)  -\frac{p}{2} + \frac{1}{2}\log \left ( \frac{|\widetilde \Sigma_n|^{-1}}{| \text{diag}(\alpha  n V_{\theta^*})|}\right) \, .
\end{align}
To prove \eqref{eq:KL_convergenceThm2}, we will prove that both $T_1=o_{f_{0,n}(1)}$ and $T_2=o_{f_{0,n}(1)}$. 
The key step to establishing these results is the asymptotic representation result of Lemma \ref{lemma3}. It shows that, under Assumptions \ref{asn:A1} and \ref{asn:A2}, for any sequence $(\mu_n, \Sigma_n)$---possibly dependent on the data---that is bounded in $f_{0,n}$-probability, we have: 
\begin{equation*}
    \mathcal{K}( q(\cdot \, | \,\mu_n, \, \Sigma_n  )   \: ||\:  \pi_{n,\alpha}(\cdot  \,| \, X^n) )= \mathcal{K} \left ( q(\cdot \, | \,\mu_n, \, \Sigma_n  )   \: ||\:  \phi(\cdot \, | \, \widehat{\theta}_{\textrm{ML\,-\,$\mathcal{F}_n$}}, \, V_{\theta^*}^{-1} / (\alpha n)  ) \right) + o_{f_{0,n}}(1) \, . 
\end{equation*}

This means that the KL divergence between any normal density $q(\cdot | \mu_n, \Sigma_n)$ and the $\alpha$-posterior $\pi_{n,\alpha}(\cdot | X^n)$ is eventually close to the KL divergence between the same density and the $\alpha$-posterior's total variation limit (which we have characterized in Theorem \ref{thm:BVMT-alpha}).

We use two intermediate steps to relate $(\widetilde{\mu}_n, \widetilde{\Sigma}_n)$ to $(\widehat{\theta}_{\textrm{ML}-\mathcal{F}_n}, \textrm{diag}(V_{\theta^*})^{-1}/(\alpha n))$. 

\noindent \textbf{Claim 1. } We start by showing that $T_1=o_{f_{0,n}}(1)$. Since $\widetilde{\mu}_n$ and $\widetilde{\Sigma}_n$ are the parameters that solve the variational approximation in \eqref{eqn:GMF-approx}, they are thus the parameters that minimize the KL divergence between the Gaussian mean-field family and the $\alpha$-posterior. It follows that for every $n$,
\begin{equation}
\label{eq:claim1_bound}
  \mathcal{K} (  q(\cdot \, | \, \widetilde{\mu}_n, \, \widetilde{\Sigma}_n  )    \: || \: \pi_{n,\alpha}(\cdot \, | \, X^n)  ) \le \mathcal{K} ( q(\cdot \,|\, \widehat{\theta}_{\textrm{ML\,-\,$\mathcal{F}_n$}}, \widetilde \Sigma_n)  \: || \: \pi_{n,\alpha}(\cdot \, | \, X^n)  ) \, . 
\end{equation}
Using Lemma \ref{lemma3}, we can evaluate each of the KL divergences above up to an $o_{f_{0,n}}(1)$ term using the asymptotic Gaussian limit of $\pi_{n,\alpha}(\cdot \, | \, X^n)$ given in Thm~\ref{thm:BVMT-alpha}. Indeed, since both $(\sqrt{n}(\widetilde{\mu}_n - \theta^*) , \, n \widetilde{\Sigma}_n)$ and $(\sqrt{n}(\widehat{\theta}_{\textrm{ML\,-\,$\mathcal{F}_n$}} - \theta^*) , \, n \widetilde{\Sigma}_n)$ are bounded in $f_{0,n}$-probability, we can apply Lemma \ref{lemma3} to find
\begin{align*}
  \mathcal{K} (  q(\cdot \, | \, \widetilde{\mu}_n, \, \widetilde{\Sigma}_n  )    \: || \: \pi_{n,\alpha}(\cdot \, | \, X^n)  ) & = \mathcal{K} (  q(\cdot \, | \, \widetilde{\mu}_n, \, \widetilde{\Sigma}_n  )    \: || \: \phi(\cdot \, | \, \widehat{\theta}_{\textrm{ML\,-\,$\mathcal{F}_n$}}, \, V_{\theta^*}^{-1} / (\alpha n)  )  )  + o_{f_{0,n}}(1)\, , \\
    \mathcal{K} (  q(\cdot \,|\, \widehat{\theta}_{\textrm{ML\,-\,$\mathcal{F}_n$}}, \widetilde \Sigma_n)   \: || \: \pi_{n,\alpha}(\cdot \, | \, X^n)  ) & = \mathcal{K} (  q(\cdot \,|\, \widehat{\theta}_{\textrm{ML\,-\,$\mathcal{F}_n$}}, \widetilde \Sigma_n)   \: || \: \phi(\cdot \, | \, \widehat{\theta}_{\textrm{ML\,-\,$\mathcal{F}_n$}}, \, V_{\theta^*}^{-1} / (\alpha n)  )  )  + o_{f_{0,n}}(1)\, , \\
\end{align*}
and plugging the above into \eqref{eq:claim1_bound} gives
\begin{equation}
\label{eq_claim1_KL}
\begin{split}
    \mathcal{K} & \left ( q(\cdot \, | \,\widetilde{\mu}_n, \, \widetilde{\Sigma}_n  )   \: ||\:  \phi(\cdot \, | \, \widehat{\theta}_{\textrm{ML\,-\,$\mathcal{F}_n$}}, \, V_{\theta^*}^{-1} / (\alpha n)  ) \right)  \\
    & \leq  \mathcal{K} \left ( q(\cdot \, | \,\widehat{\theta}_{\textrm{ML\,-\,$\mathcal{F}_n$}}, \, \widetilde{\Sigma}_n  )   \: ||\:  \phi(\cdot \, | \, \widehat{\theta}_{\textrm{ML\,-\,$\mathcal{F}_n$}}, \, V_{\theta^*}^{-1} / (\alpha n)  ) \right) + o_{f_{0,n}}(1).
\end{split}
\end{equation}
 Therefore, the result follows from \eqref{eq:KLnormals} and  \eqref{eq_claim1_KL} that
\begin{align}
&  \nonumber T_1 = \frac{1}{2}  (\widehat{\theta}_{\textrm{ML\,-\,$\mathcal{F}_n$}} - \widetilde\mu_n)^\top (\alpha n V_{\theta^*})(\widehat{\theta}_{\textrm{ML\,-\,$\mathcal{F}_n$}} - \widetilde\mu_n) \\
  \nonumber& =  \mathcal{K}  \left ( q(\cdot \, | \,\widetilde{\mu}_n, \, \widetilde{\Sigma}_n  )   \: ||\:  \phi(\cdot \, | \, \widehat{\theta}_{\textrm{ML\,-\,$\mathcal{F}_n$}}, \,  V_{\theta^*}^{-1} / (\alpha n)    ) \right)  -   \mathcal{K} \left ( q(\cdot \, | \,\widehat{\theta}_{\textrm{ML\,-\,$\mathcal{F}_n$}}, \, \widetilde{\Sigma}_n  )   \: ||\:  \phi(\cdot \, | \, \widehat{\theta}_{\textrm{ML\,-\,$\mathcal{F}_n$}}, \,  V_{\theta^*}^{-1} / (\alpha n)   ) \right)\\
  &\leq  o_{f_{0,n}}(1) \, .\label{eqn:part1_thm2}
\end{align}

\noindent \textbf{Claim 2.} We now show that $T_2=o_{f_{0,n}}(1)$ by relating $\widetilde{\Sigma}_n$ to $\diag(V_{\theta^*})^{-1} / (\alpha n )$.
The optimality of $\widetilde{\mu}_n$ and $\widetilde{\Sigma}_n$ defined by \eqref{eqn:GMF-approx} once again implies that for every $n$:
\begin{equation}
       \label{eq_claim2_KL}
\mathcal{K} ( q(\cdot \,|\, \widetilde \mu_n, \widetilde \Sigma_n )\: || \: \pi_{n,\alpha}(\cdot \, | \, X^n)  ) \le \mathcal{K} ( q(\cdot \,|\, \widehat{\theta}_{\textrm{ML\,-\,$\mathcal{F}_n$}}, \text{diag}(V_{\theta^*})^{-1}/(\alpha n) )  \: || \: \pi_{n,\alpha}(\cdot \, | \, X^n)  ). 
\end{equation} 
As in the work of Claim 1, this inequality and Lemma \ref{lemma3} imply that
\begin{equation}
   \begin{split}
   \label{eq:claim2_eq2}
       &\mathcal{K} ( q(\cdot \,|\, \widetilde \mu_n, \widetilde \Sigma_n)\: || \:   \phi(\cdot \, | \, \widehat{\theta}_{\textrm{ML\,-\,$\mathcal{F}_n$}}, \, V_{\theta^*}^{-1} / (\alpha n)  ) ) \\
   &\le \mathcal{K} ( q(\cdot \,|\, \widehat{\theta}_{\textrm{ML\,-\,$\mathcal{F}_n$}}, \text{diag}(V_{\theta^*})^{-1}/(\alpha n) )  \: || \:  \phi(\cdot \, | \, \widehat{\theta}_{\textrm{ML\,-\,$\mathcal{F}_n$}}, \, V_{\theta^*}^{-1} / (\alpha n)  ) )+o_{f_{0,n}}(1).
   \end{split} 
\end{equation}
Next, using the inequality in \eqref{eq:claim2_eq2}, applying \eqref{eq:KLnormals} to each term, and
noting that \eqref{eqn:part1_thm2} implies  $\frac{1}{2}  (\widehat{\theta}_{\textrm{ML\,-\,$\mathcal{F}_n$}} - \widetilde\mu_n)^\top (\alpha n V_{\theta^*})(\widehat{\theta}_{\textrm{ML\,-\,$\mathcal{F}_n$}} - \widetilde\mu_n) \leq  o_{f_{0,n}}(1)$, we obtain
\begin{align*}
&\frac{1}{2} \left[ \tr(\alpha n V_{\theta^*} \widetilde \Sigma_n) - p + \log \left(\frac{|V_{\theta^*}^{-1}/(\alpha n )|}{|\widetilde \Sigma_n|} \right) \right] \\
&\leq   \frac{1}{2} \left[ \tr( \alpha n V_{\theta^*} \text{diag}(V_{\theta^*})^{-1}/(\alpha n) ) - p + \log \left(\frac{|V_{\theta^*}^{-1}/(\alpha n )|}{|\text{diag}(V_{\theta^*})^{-1}/(\alpha n)|} \right)  \right]+ o_{f_{0,n}}(1) \, .
\end{align*}
Further noting that $ \tr( V_{\theta^*} \diag(V_{\theta^*})^{-1} ) = p$, we see that the above inequality is equivalent to
\begin{equation}
\begin{split}
\label{eqn:part2_thm2}
 &\frac{1}{2} \left[ \tr(\alpha n V_{\theta^*} \widetilde \Sigma_n) - p + \log \left(\frac{|V_{\theta^*}^{-1}/(\alpha n )|}{|\widetilde \Sigma_n|} \right) - \log \left(\frac{|V_{\theta^*}^{-1}/(\alpha n )|}{|\text{diag}(V_{\theta^*})^{-1}/(\alpha n)|} \right)  \right]    \leq   o_{f_{0,n}}(1) \, .
 \end{split}
\end{equation}
Since $\widetilde{\Sigma}_n$ is diagonal,
\[ \tr(\alpha n V_{\theta^*} \widetilde \Sigma_n) = \tr( \textrm{diag}(\alpha n V_{\theta^*}) \widetilde \Sigma_n), \]
and consequently, the left-hand side of \eqref{eqn:part2_thm2} equals $T_2$, which has been defined in \eqref{eq:T2}. Moreover, since the term $T_2$ is nonnegative, as it equals the KL divergence between two normals with the same mean, but variances $\widetilde{\Sigma}_n$ and $\textrm{diag}(V_{\theta^*})^{-1}/\alpha n$, we conclude that $T_2=o_{f_{0,n}}(1)$. 

%%%%%%%%%%%%%%  Proof of Theorem 3 %%%%%%%%%%%%%%%%%%%
\subsection{Proof of Theorem 3}\label{proof:thm3}
We note that $r_n^*$ and $\tilde{r}_n^*$ defined in \eqref{eq:rn_star} and \eqref{eq:tilde_rn_star} are both expectations involving KL divergences of two multivariate Gaussian distributions, hence
\eqref{eq:KLnormals} allows us to compute $r_n^*(\alpha)$ and $\widetilde{r}^*(\alpha)$ explicitly as  
\begin{align*}
    r_n^*(\alpha) = \frac{1}{2} \Big(   \alpha A_n(V_{\theta^*}) - p \log(\alpha) + B_n(V_{\theta^*})  \Big), \quad  \widetilde{r}_n^*(\alpha) = \frac{1}{2} \Big(   \alpha A_n(\widetilde{V}_{\theta^*}) - p \log(\alpha) + B_n(\widetilde{V}_{\theta^*})  \Big) \,,
\end{align*}
where 
\begin{align*}
    A_n(\Sigma) &\equiv \epsilon_n \tr(\Sigma \Omega)  + (1-\epsilon_n) \tr(\Sigma V_{\theta^*}^{-1}) +  n \epsilon_n (\widehat{\theta}_{\textrm{ML\,-\,$\mathcal{F}_n$}}-\widehat{\theta}_{\textrm{ML}})^{\top} \Sigma(\widehat{\theta}_{\textrm{ML\,-\,$\mathcal{F}_n$}}-\widehat{\theta}_{\textrm{ML}}) \, ,\\
   B_n(\Sigma) &\equiv -p + \epsilon_n \log ( |\Omega|^{-1}|\Sigma^{-1}|) + (1-\epsilon_n) \log ( |V_{\theta^*}^{-1}|^{-1}|\Sigma^{-1}|) \,.
\end{align*}
Using this notation, we see that $r_n^*$ and $\widetilde{r}_n^*$ are convex functions on $\alpha$. This implies that first order conditions pin-down the optimal $\alpha_n^*$ and $\widetilde \alpha_n^*$ defined in \eqref{eqn:alpha_robust}. These are equal to
\begin{equation}\label{eqn:aux0_thm3}
    \alpha_n^* = \frac{p}{A_n(V_{\theta^*})} \quad \text{and} \quad \widetilde \alpha_n^* = \frac{p}{A_n(\widetilde{V}_{\theta^*})} \, .
\end{equation} 
By assumption we have $\widehat{\theta}_{\textrm{ML}}\overset{p}{\rightarrow} \theta_0$ and $n \epsilon_n \to \varepsilon \in (0, \infty)$. In addition, we know that $\widehat{\theta}_{\textrm{ML\,-\,$\mathcal{F}_n$}} \overset{p}{\rightarrow} \theta^*$. It follows that
$ A_n(\Sigma) \overset{p}{\rightarrow} \tr(\Sigma V_{\theta^*}^{-1}) + \varepsilon (\theta^* - \theta_0)^{\top} \Sigma (\theta^* - \theta_0).$
Denote $\alpha^*$ and $\widetilde{\alpha}^*$ the limits in $f_{0,n}$-probability of $\alpha_n^*$ and $\widetilde \alpha_n^*$. These limits can be computed replacing $\Sigma$ in the expressions above with $V_{\theta^*}$ or $\widetilde{V}_{\theta^*}$ and taking limit of \eqref{eqn:aux0_thm3}. This implies
\begin{eqnarray*}
\alpha^*_n  &\overset{p}{\rightarrow} \alpha^* \equiv \frac{p}{p + \varepsilon (\theta_0-\theta^*)^\top V_{\theta^*}(\theta_0-\theta^*)   } \, ,   \qquad \widetilde{\alpha}^*_n  \overset{p}{\rightarrow} \widetilde{\alpha}^* \equiv \frac{p}{\textrm{tr}(  \widetilde{V}_{\theta^*} V^{-1}_{\theta^*}) +  \varepsilon (\theta_0-\theta^*)^\top \widetilde{V}_{\theta^*}(\theta_0-\theta^*) } \,.
\end{eqnarray*}
To conclude $\widetilde \alpha^*<1$ is sufficient to prove that $\textrm{tr}(  \widetilde{V}_{\theta^*} V^{-1}_{\theta^*}) \ge p$.

Recall that the trace of a matrix is the sum of the eigenvalues and the determinant is the product. Applying the Arithmetic Mean-Geometric Mean inequality, 
we have
$$\frac{1}{p}\tr\left(  \widetilde{V}_{\theta^*} V^{-1}_{\theta^*}\right) \ge \left( |\widetilde{V}_{\theta^*} V^{-1}_{\theta^*}| \right)^{1/p}.$$
Then since $|\widetilde{V}_{\theta^*} V^{-1}_{\theta^*}|  = |\widetilde{V}_{\theta^*}| |V^{-1}_{\theta^*}| = |\widetilde{V}_{\theta^*}| |V_{\theta^*}|^{-1}$, it will be sufficient to prove that $|\widetilde{V}_{\theta^*}| \ge |V_{\theta^*}|$, where $\widetilde{V}_{\theta^*} = \text{diag}(V_{\theta^*})$. Because $V_{\theta^*}$ is a semi-definite matrix, this is exactly Hadamard's inequality (see Theorem 7.8.1 in \cite{HornJohnson:2012}).

%%%%%%%%%%%%%%%% 
\section{Technical work for the illustrative example}\label{section:proof_example}

\subsection{Verification of $\alpha$-posterior concentration}\label{section:B1}
We first want to prove that the $\alpha$-posterior concentrates at the rate $\sqrt{n}$ around $\theta^*$ as defined in \eqref{eqn:concentration}. In other words need to show that for every sequence $r_n \to \infty$, we have
\begin{equation}\label{eqn:aux0_example}
   \mathbb{E}_{f_{0,n}} \left[ \, \mathbb{P}_{\pi_{n,\alpha} (\cdot \, | \, X^n)} \left( \| \sqrt{n} \, (\theta-\theta^*) \| > r_n   \right) \right] \to 0 \, . 
\end{equation}

\textbf{Step 1:} Compute an upper bound for the probability inside the brackets using Markov's inequality. 

Note that in our illustrative example, we have $\pi_{n,\alpha} (\theta \, | \, X^n) \sim \mathcal{N}(\mu_{n,\alpha},\: \Sigma_{n,\alpha})$, where $\mu_{n,\alpha} $ and $\Sigma_{n,\alpha}$ were defined in \eqref{eqn:alpha-mean-linref} and \eqref{eqn:alpha-variance-linref}. Therefore, by Markov's inequality and the normal distribution of the $\alpha$-posterior we have 
\begin{align}
\mathbb{P}_{\pi_{n,\alpha} (\cdot \, | \, X^n)} \left( \| \theta-\theta^* \|^2 > r_n^2/n   \right) & \leq \frac{n}{r_n^2} \mathbb{E}_{\pi_{n,\alpha} (\cdot \, | \, X^n)} \left[ \| \theta-\theta^* \|^2  \right] = \frac{n}{r_n^2} \left[ \| \mu_{n,\alpha} - \theta^*\|^2 + \text{tr}(\Sigma_{n,\alpha}) \right] \, ,\label{eqn:aux1_example}
\end{align}
which defines the upper bound for the probability inside the brackets.

\textbf{Step 2:} Conclude that the expected value of the probability goes to zero. 

Lemma \ref{lemma4} in Appendix \ref{section:technical} implies that the sequence $(\sqrt{n}(\mu_{n,\alpha}-\theta^*),\: n \Sigma_{n,\alpha})$ is bounded in $f_{0,n}$-probability. It follows that both $\|\sqrt{n}(\mu_{n,\alpha} - \theta^*)\|^2$ and $\text{tr}(n \Sigma_{n, \alpha})$ are bounded in $f_{0,n}$-probability. This means that for every $\epsilon>0$, there exists an $M_\epsilon>0$ such that 
\begin{equation}\label{eqn:aux2_example}
    \mathbb{P}_{f_{0,n}} \left( A_\epsilon \right) \le \epsilon \,,
\end{equation}
where $A_{\epsilon} = \{\|\sqrt{n}(\mu_{n,\alpha} - \theta^*)\|^2 + \text{tr}(n \Sigma_{n, \alpha}) > M_\epsilon\}$. By linearity of the expectation, we have that for any sequence $r_n$ and for any $\epsilon>0$:
\begin{align}
\nonumber        \mathbb{E}_{f_{0,n}} &\left[ \, \mathbb{P}_{\pi_{n,\alpha} (\cdot \, | \, X^n)} \left( \| \sqrt{n} \, (\theta-\theta^*) \| > r_n   \right) \right] \\
     &\le \mathbb{E}_{f_{0,n}} \left[ \, \mathbb{P}_{\pi_{n,\alpha} (\cdot \, | \, X^n)} \left( \| \sqrt{n} \, (\theta-\theta^*) \| > r_n   \right) 1\{A_\epsilon^c\} \right] +  \mathbb{E}_{f_{0,n}} \left[ 1\{ A_\epsilon \} \right] \, ,
      \label{eqn:aux3_example}  
\end{align}
where $1\{A_\epsilon\}$ is the indicator function of the event $A_\epsilon$.
The first term in  \eqref{eqn:aux3_example} can be  bounded using \eqref{eqn:aux1_example} and the definition of $A_\epsilon$, leading to 
\begin{align*}
    \mathbb{E}_{f_{0,n}} & \left[ \, \mathbb{P}_{\pi_{n,\alpha} (\cdot \, | \, X^n)} \left( \| \sqrt{n} \, (\theta-\theta^*) \| > r_n   \right) 1\{A_\epsilon^c\} \right] \\
    &\leq \mathbb{E}_{f_{0,n}} \left[ \frac{1}{r_n^2} \left[ \| \sqrt{n}( \mu_{n,\alpha} - \theta^*)\|^2 + \text{tr}(n\Sigma_{n,\alpha}) \right] 1\{A_\epsilon^c\} \right]\leq \frac{M_\epsilon}{r_n^2} \, .
\end{align*}
Using \eqref{eqn:aux2_example}, we see that the second term in  \eqref{eqn:aux3_example} is smaller than $\epsilon$. Hence, we conclude that
$$ \mathbb{E}_{f_{0,n}} [ \, \mathbb{P}_{\pi_{n,\alpha} (\cdot \, | \, X^n)} \left( \| \sqrt{n} \, (\theta-\theta^*) \| > r_n   \right) ] \le \frac{M_\epsilon}{r_n^2} + \epsilon \, ,$$
which is sufficiently small since $\epsilon>0$ was arbitrary, $M_{\epsilon}$ is constant, and $r_n \to \infty$. This verifies \eqref{eqn:aux0_example}.

\subsection{KL distance for Theorem 1}\label{section:B2}
In our illustrative example, we can compute the KL distance between the $\alpha$-posterior distribution $\pi_{n,\alpha}$ and the distribution defined in \eqref{eqn:limit_regression} since both distribution are multivariate normal. Using \eqref{eq:KLnormals}, we find the KL divergence ot equal
$$ \frac{1}{2} \left( -p + \log \left( \frac{|V_{\theta^*} \alpha n|^{-1}}{|\Sigma_{n,\alpha}|} \right) + \text{tr}(V_{\theta^*} \alpha n\Sigma_{n,\alpha}) + (\mu_{n,\alpha}-\widehat{\theta}_{ML})^{\top} V_{\theta^*} \alpha n (\mu_{n,\alpha}-\widehat{\theta}_{ML})  \right).$$
Lemma \ref{lemma4} implies that the previous expression converges to 0 in $f_{0,n}$-probability. This means that the KL distance between $\pi_{n,\alpha}$ and the distribution in \eqref{eqn:limit_regression} goes to zero in $f_{0,n}$-probability.

\subsection{What if $\alpha$ goes to zero very quickly?}\label{section:B3}
Suppose that the sequence of $\alpha_n$ satisfies $n \alpha_n \to \alpha_0>0$. This implies that, in $f_{0,n}$-probability,
\begin{align*}
   \mu_{n, \alpha_n} \to \left[ \mathbb{E}[W_iW_i^\top] + \frac{\Sigma_{\pi} }{\alpha_0 } \right]^{-1} \left[ \frac{ \Sigma_{\pi} \mu_\pi}{\alpha_0 } +  \mathbb{E}[W_iW_i^\top] \theta^* \right], \qquad \Sigma_{n,\alpha_n} \to & \frac{\sigma^2_u}{\alpha_0}\left[ \mathbb{E}[W_i W_i^\top] + \frac{\Sigma_{\pi}}{\alpha_0 }  \right]^{-1} \,,
\end{align*}
Using these different limits, we show that the total variation distance between the  $\alpha_n$-posterior distribution $\pi_{n,\alpha_n}$ and the distribution in \eqref{eqn:limit_regression} is bounded away from zero. We show this using that the square of the Hellinger distance is a lower bound for the total variation distance.

In our illustrative example, the $\alpha_n$-posterior distribution $\pi_{n,\alpha_n}$ and the distribution in \eqref{eqn:limit_regression} are both multivariate normal distributions. Then, we can compute the square of the Hellinger distance between these distributions. Using Lemma B.1 part ii) of \cite{ghosal2017}, we obtain
\begin{equation*}
    \begin{split}
    1 - &\frac{|\Sigma_{n,\alpha_n}|^{1/4}|(V_{\theta^*} \alpha_n n)^{-1}|^{1/4}}{|(\Sigma_{n,\alpha_n}+(V_{\theta^*} \alpha_n n)^{-1})/2|^{1/2}}  \exp \left( -\frac{1}{8} (\mu_{n,\alpha_n}-\widehat{\theta}_{ML})^{\top} \frac{\Sigma_{n,\alpha_n}+(V_{\theta^*} \alpha_n n)^{-1}}{2} (\mu_{n,\alpha_n}-\widehat{\theta}_{ML}) \right)    \, ,
    \end{split}
\end{equation*}
 which converge in $f_{0,n}$-probability to a positive number. To verify this, notice that $\Sigma_{n,\alpha_n}$ and $(V_{\theta^*} \alpha_n n)^{-1}$ converge to different limits, which guarantees 
$$ \lim_{n \to \infty} \frac{|\Sigma_{n,\alpha_n}|^{1/4}|(V_{\theta^*} \alpha_n n)^{-1}|^{1/4}}{|(\Sigma_{n,\alpha_n}+(V_{\theta^*} \alpha_n n)^{-1})/2|^{1/2}} < 1 \, ,$$ 
where the limit is taken in $f_{0,n}$-probability and the inequality follows by applying the Arithmetic Mean-Geometric Mean inequality.

\subsection{Verification of Assumption 2}\label{section:B4}

In our illustrative example:
\begin{align*}
    \pi(\theta) &\sim \mathcal{N}(\mu_{\pi}, \sigma_u^2 \Sigma_{\pi}^{-1}), \quad \text{ and } \quad 
    R_n(h) = h^{\top} Q_n \Delta_{n,\theta^*} - \frac{1}{2} h^{\top} Q_n h \, ,
\end{align*}
where $\Delta_{n,\theta^*} = \sqrt{n}(\widehat{\theta}_{ML}-\theta^*)$ and 
$$ Q_n \equiv  \frac{\sum_{i=1}^n W_iW_i^\top}{n \sigma_u^2} - V_{\theta^*} \, .$$
$Q_n$ converges to zero in $f_{0,n}$-probability in our illustrative example since $V_{\theta^*} = \mathbb{E}[W_iW_i^\top]/\sigma_u^2$.

Let us take a sequence $(\mu_n,\Sigma_n)$ such that $(\sqrt{n}(\mu_n-\theta^*), n\Sigma_n)$ is bounded in $f_{0,n}$-probability. Then, equation \eqref{eq:asn2.1} in Assumption \ref{asn:A2} becomes
\begin{equation}
\begin{aligned}
       & \int \phi(h\,|\, \sqrt{n}(\mu_n-\theta^*), n\Sigma_n) \left( -\frac{1}{2} \frac{h^{\top}}{\sqrt{n}}\frac{\Sigma_\pi}{\sigma_u^2} \frac{h}{\sqrt{n}} + \frac{h^{\top}}{\sqrt{n}} \frac{\Sigma_\pi}{\sigma_u^2} (\mu_\pi-\theta^*) \right) dh\\ \noalign{\vskip10pt}
    &= -\frac{1}{2n} \overline \mu_n^{\top} \frac{\Sigma_\pi}{\sigma_u^2} \overline \mu_n - \frac{1}{2n} \text{tr} \left( n \Sigma_n \frac{\Sigma_\pi}{\sigma_u^2} \right) + \frac{1}{\sqrt{n}}\overline \mu_n^{\top} \frac{\Sigma_\pi}{\sigma_u^2} (\mu_\pi-\theta^*) \, ,
    \end{aligned}
    \label{eq:B4}
\end{equation} 
where $\overline \mu_n = \sqrt{n}(\mu_n-\theta^*)$. By assumption, the sequence $(\overline \mu_n, n\Sigma_n)$ is bounded in $f_{0,n}$-probability. This implies that \eqref{eq:B4} goes to zero in $f_{0,n}$-probability.

Equation \eqref{eq:asn2.2} in Assumption \ref{asn:A2} can be computed explicitly as
\begin{equation*}
    \in \phi(h| \sqrt{n}(\mu_n-\theta^*), n\Sigma_n) \Big( h^{\top} Q_n \Delta_{n,\theta^*} - \frac{1}{2} h^{\top} Q_n h \Big) dh =\overline \mu_n^{\top} Q_n \Delta_{n,\theta^*} -\frac{1}{2}  \mu_n^{\top} Q_n \overline \mu_n - \frac{1}{2} \text{tr}(Q_n n \Sigma_n) \,.
\end{equation*} 
By assumption, the sequence $(\overline \mu_n, n\Sigma_n)$ is bounded in $f_{0,n}$-probability. Since $Q_n$ converge to zero in $f_{0,n}$-probability, the expression above goes to zero in $f_{0,n}$-probability. 

\subsection{KL distance for Theorem 2}
Lemma \ref{lemma4} shows that in our illustrative example, the sequence $(\mu_{n,\alpha}, \Sigma_{n,\alpha})$ verifies that $(\sqrt{n}(\mu_{n,\alpha}-\theta^*),\: n \Sigma_{n,\alpha})$ is bounded in $f_{0,n}$-probability. Then, we can apply Theorem \ref{thm:BVMT-alpha-var}. The proof presented in Section \ref{proof:thm2} shows that, in $f_{0,n}$-probability,
$$ \mathcal{K} ( \widetilde \pi_{n,\alpha}(\cdot \,|\, X^n )  \: || \: \phi(\cdot \, | \, \widehat{\theta}_{\textrm{ML\,-\,$\mathcal{F}_n$}}, \, \textrm{diag}(V_{\theta^*})^{-1} / (\alpha n)  )  ) \to 0 \, .$$
This means that the Bernstein-von Mises Theorem holds for the variational approximation to the $\alpha$-posterior in KL divergence.

\subsection{About the optimal $\alpha$ for robustness}\label{section:B6}

Let us recall the definition of $r_n(\alpha)$ presented in Section \ref{section:Robustness} in equation \eqref{eqn:Ratio}:
\begin{equation*}  
\begin{split}
    r_n(\alpha) & =  \epsilon_n \mathcal{K} \left( \phi(\cdot\,|\, \nu_n^*,\: \Omega_n^*) \: || \: \phi(\cdot\,|\, \mu_{n,\alpha},\: \Sigma_{n,\alpha})\right)  +  (1-\epsilon_n) \mathcal{K} \left( \phi(\cdot\,|\,\mu_{n,1},\: \Sigma_{n,1} )  \: || \: \phi(\cdot\,|\, \mu_{n,\alpha},\: \Sigma_{n,\alpha}  ) \right),
\end{split}
\end{equation*}
where $\nu_n^*$ is the true posterior mean and $\Omega_n^*$ is the true posterior covariance matrix. The expression above is equal to
\begin{equation*} 
\begin{split}
    r_n(\alpha) =& \frac{1}{2}\Big\{  \epsilon_n \tr(\Sigma_{n,\alpha}^{-1} \Omega_n^* ) + \alpha (1-\epsilon_n) \text{tr}((\alpha n \Sigma_{n,\alpha})^{-1} n \Sigma_{n,1})\\
    &+ \alpha n \epsilon_n (\mu_{n,\alpha}-\nu_n^*)^\top (\alpha n\Sigma_{n, \alpha})^{-1} (\mu_{n,\alpha}-\nu_n^*)\\
    &+ \alpha n (1- \epsilon_n) (\mu_{n,1}-\mu_{n,\alpha})^\top (\alpha n \Sigma_{n, \alpha})^{-1} (\mu_{n,1}-\mu_{n,\alpha})\\
     &-p \log(\alpha) - p + \epsilon_n \log\left(\frac{|\Sigma_{n,\alpha}|}{|\Omega_n^*|}\right) + (1-\epsilon_n)  \log\left(\frac{|\Sigma_{n,\alpha}|}{|\Sigma_{n,1}|}\right)\Big\} \, .
\end{split}
\end{equation*}

Notice that $\Sigma_{n,\alpha}^{-1}\Omega_n^* \to \alpha V_{\theta^*} \Omega$ in $f_{0,n}$-probability, since it can be proved that $n \Omega_n^* \to \Omega$ in the well-specified model, for some definite positive matrix $\Omega$. Lemma \ref{lemma4} implies that $(\alpha n \Sigma_{n,\alpha})^{-1} n \Sigma_{n,1} \to \mathbb{I}_p$ in $f_{0,n}$-probability. Moreover, we have that $n(\mu_{n,1}-\mu_{n,\alpha})$ is bounded in $f_{0,n}$-probability, which implies that
$ (\mu_{n,1}-\mu_{n,\alpha})^\top \Sigma_{n, \alpha} (\mu_{n,1}-\mu_{n,\alpha}) \to 0$
in $f_{0,n}$-probability. Since $n \epsilon_n \to \varepsilon$, we conclude that, in $f_{0,n}$-probability,
$ r_n(\alpha) \to r_\infty(\alpha)$ where $r_\infty(\alpha) \equiv \frac{1}{2} ( \alpha p + \alpha \varepsilon (\theta^*-\theta_0)^{\top} V_{\theta^*} (\theta^*-\theta_0) - p\log (\alpha) -p)$.

Using the notation introduced in the proof of Theorem \ref{thm:limit_r} in Section \ref{proof:thm3}, we have 
$$r_n^*(\alpha) =  \frac{1}{2} \Big(   \alpha\: A_n(V_{\theta^*}) - p \log(\alpha) + B_n(V_{\theta}^*)  \Big) \,,$$
where 
\begin{align*}
    A_n(\Sigma) \equiv& \epsilon_n \tr(\Sigma \Omega)  + (1-\epsilon_n) \tr(\Sigma V_{\theta^*}^{-1}) +  n \epsilon_n (\widehat{\theta}_{\textrm{ML\,-\,$\mathcal{F}_n$}}-\widehat{\theta}_{\textrm{ML}})^{\top} \Sigma(\widehat{\theta}_{\textrm{ML\,-\,$\mathcal{F}_n$}}-\widehat{\theta}_{\textrm{ML}}) \, ,\\
    B_n(\Sigma) \equiv& -p + \epsilon_n \log ( |\Omega^{-1}||\Sigma|^{-1}) + (1-\epsilon_n) \log ( |V_{\theta^*}||\Sigma|^{-1}) \, .
\end{align*}
Notice that $A_n(V_{\theta^*}) \to p + \varepsilon (\theta^*-\theta_0)^{\top} V_{\theta^*}(\theta^*-\theta_0)$ and $B_n(V_{\theta^*}) \to -p$ in $f_{0,n}$-probability. This implies that
$$r_n^*(\alpha) \to r_\infty(\alpha) = \frac{1}{2} \left( \alpha p + \alpha \varepsilon (\theta^*-\theta_0)^{\top} V_{\theta^*} (\theta^*-\theta_0) - p\log (\alpha) -p \right) \, .$$
Then, we can conclude that
$r_n(\alpha) - r_n^*(\alpha) \to 0$
in $f_{0,n}$-probability. In particular, for $\alpha = \alpha^*$ defined in Theorem \ref{thm:limit_r} and any $\alpha' \neq \alpha^*$, we have that $r_n(\alpha^*)$ is close to $r_\infty(\alpha^*)$ and $r_n(\alpha')$ is close to $r_\infty(\alpha')$. Since $r_\infty(\alpha^*)< r_\infty(\alpha')$ by definition of $\alpha^*$, it follows that for large $n$, $r_n(\alpha^*) < r_n(\alpha')$.

%\newpage

\section{Technical Lemmas} \label{section:technical} 

\begin{lemma}\label{lemma1}
Consider sequences of densities $\phi_n$ and $\psi_n$. For a given compact set $K \subset \mathbb{R}^p$, suppose that the densities $\psi_n$ and $\phi_n$ are positive on $K$. Then, 
$$ d_{\textrm{TV}}\left( \psi_n, \, \phi_n \right) \le \sup_{g,h \, \in \, K} f_n(g,h) +  \int_{\mathbb{R}^p \setminus K} \psi_n(h) \, dh \, + \, \int_{\mathbb{R}^p \setminus K} \phi_n(h) \, dh  \, , $$
where we have defined the function
\begin{equation} 
f_n(g,h) = \Big \{ 1 - \frac{\phi_n(h)}{\psi_n(h)} \, \frac{\psi_n(g)}{\phi_n(g)}   \Big \}^+ \,.
\label{eq:fn_def}
\end{equation}
\end{lemma}

% %%%% 
\noindent \textbf{Proof.} First, denote $a_n = \left\{\int_K \psi_n(g) \, dg\right\}^{-1}$ and $b_n = \left\{\int_K \phi_n(g) \, dg\right\}^{-1}$. Notice that both are well defined since $\phi$ and $\psi$ are assumed positive on $K$. We will assume throughout that $a_n \geq b_n$, without loss of generality.

First, by definition, $d_{\textrm{TV}}\left( \psi_n, \phi_n \right) = \frac{1}{2} \int |\psi_n(h) - \phi_n(h)| \, dh$. Then since $|x| = 2 \{x\}^+ - x$, it follows that the total variation is equal to
 \begin{equation}
 \begin{split}
     d_{\textrm{TV}}\left( \psi_n, \, \phi_n \right) &= \frac{1}{2} \int |\psi_n(h) - \phi_n(h)| \, dh \\
     &=  \int_{\mathbb{R}^p} \Big \{ \psi_n(h) - \phi_n(h) \Big \}^+ dh + \frac{1}{2} \int_{\mathbb{R}^p} (\psi_n(h) - \phi_n(h)) \, dh \\
     &=    \int_{\mathbb{R}^p} \Big \{ \psi_n(h) - \phi_n(h) \Big \}^+ \, dh \\
 &=   \int_{K} \Big \{ \psi_n(h) - \phi_n(h) \Big \}^+ \, dh  +  \int_{\mathbb{R}^p \setminus K} \Big \{ \psi_n(h) - \phi_n(h) \Big \}^+ \, dh \,  .
     \label{eq:lemma_bound1}
 \end{split}
 \end{equation}
Now, since $\psi_n, \phi_n$ are non-negative on all of $\mathbb{R}^p$, it follows that $\{\psi_n(h)-\phi_n(h)\}^+ \le \psi_n(h) + \phi_n(h)$. This provides a bound for the second term on the right side of \eqref{eq:lemma_bound1}:
 \begin{equation}
 \begin{split}
 \label{eq:lemma_bound_I2}
 \int_{\mathbb{R}^p \setminus K} \Big \{ \psi_n(h) - \phi_n(h) \Big \}^+ \, dh %&\le \int_{\mathbb{R}^p \setminus K} \psi_n(h) \, dh 
 \leq  \int_{\mathbb{R}^p \setminus K} \psi_n(h) \, dh \, + \, \int_{\mathbb{R}^p \setminus K} \phi_n(h) \, dh \,.
 \end{split}
 \end{equation}

To complete the proof we show that the first term on the right side of \eqref{eq:lemma_bound1} is upper bounded by $\sup_{g,h \, \in \, K} f_n(g,h) $.
For all $h \in K$, we have the following identity,
 $$ \frac{\phi_n(h)}{\psi_n(h)} = \frac{a_n}{b_n} \, \int_K \frac{\phi_n(h)}{\psi_n(h)} \, \frac{\psi_n(g)}{\phi_n(g)} \, b_n \phi_n(g) \, dg \, .$$
 Thus,  we can rewrite the first term on the right side of \eqref{eq:lemma_bound1} as follows,
 \begin{equation}
 \begin{split}
 \int_{K} \Big \{ \psi_n(h) - \phi_n(h) \Big \}^+ \, dh &=   \int_{K} \left \{ 1 - \frac{\phi_n(h)}{\psi_n(h)} \right \}^+ \, \psi_n(h)  \, dh \\
 & = \int_{K} \Big \{ 1 -  \frac{a_n}{b_n} \, \int_K \frac{\phi_n(h)}{\psi_n(h)} \, \frac{\psi_n(g)}{\phi_n(g)} \, b_n \phi_n(g) \, dg\Big \}^+ \, \psi_n(h) \, dh  \,.
     \label{eq:lemma_I1bound1}
 \end{split}
 \end{equation}
 Now, by applying Jensen's inequality on the convex function $f(x) = \{1-x\}^+$, we have that $\left\{1 - \mathbb{E}[X]\right\}^+ \leq \mathbb{E}\left[ \left\{1 - X\right\}^+ \right]$. Applying this to the final expression above, and recalling the definition of $f_n(g,h)$ in \eqref{eq:fn_def}, we find
 \begin{equation}
 \begin{split}
 &\int_{K} \left(\Big \{ 1 -  \frac{a_n}{b_n} \, \int_K \frac{\phi_n(h)}{\psi_n(h)} \, \frac{\psi_n(g)}{\phi_n(g)} \, b_n \phi_n(g) \, dg\Big \}^+ \right) \psi_n(h) \, dh  \\
 &\leq 
 \int_{K} \left(\int_K \Big \{ 1 -  \frac{a_n}{b_n} \,  \frac{\phi_n(h)}{\psi_n(h)} \, \frac{\psi_n(g)}{\phi_n(g)}\Big \}^+ \, b_n \phi_n(g) \, dg \right)  \psi_n(h) \, dh \leq
 \int_K \int_K f_n(g,h) \, b_n \phi_n(g) \, \psi_n(h) \,  dg \,  dh \, ,
     \label{eq:lemma_I1bound2}
 \end{split}
 \end{equation}
where the final step uses that when $a_n/b_n \ge 1$, we have $\{1-(a_n/b_n) x\}^+  \leq \{1 - x\}^+$ for  $x \ge 0$. We finally note that, 
 \begin{equation*}
 \begin{split}
\int_K \int_K f_n(g,h) \, b_n \phi_n(g) \, \psi_n(h) \,  dg \,  dh &\leq   \left(\sup_{g,h \, \in \, K} f_n(g,h)\right) \left(\int_K \int_K \, b_n\phi_n(g) \, \psi_n(h) \,  dg \,  dh \right)\\
 &= \left(\int_K  \, \psi_n(h) \,  dh \right)  \left(\sup_{g,h \, \in \, K} f_n(g,h)\right) \leq \sup_{g,h \, \in \, K} f_n(g,h) \, ,
 \end{split}
 \end{equation*}
 The final equality uses that $a_n^{-1} = \int_K \psi_n(g) \, dg  \leq 1$.

\begin{lemma}\label{lemma2}
Assume  there exists a $\delta>0$ such that the prior density $\pi$ is continuous and positive on $B_{\theta^*}(\delta)$, the closed ball of radius $\delta$ around $\theta^*$ and that Assumption~\ref{asn:A1} holds. For any $\eta, \epsilon >0$, there exists a sequence $r_n \to +\infty$ and an integer $N(\eta,\epsilon) > 0$, such that for all $n > N(\eta, \epsilon)$, with $f_n(g,h)$ defined in \eqref{eq:fn_theorem_def},
$$ \mathbb{P}_{f_{0,n}}\Big ( \sup_{g,h \, \in \, \overline{B}_{\bf{0}}(r_n)} f_n(g,h) > \eta \Big ) \le \epsilon \, ,$$
where $\overline B_{\bf{0}}(r_n)$ denotes a closed ball of radius $r_n$ around $\bf{0}$.
\end{lemma}  

\noindent \textbf{Proof:}
The proof has two steps. In Step 1, we prove the claim for any fixed $r>0$, instead of a sequence $r_n$. In Step 2, we construct a sequence of $r_n$ using  equation (\ref{eq:oterm3}).

\noindent \textbf{Step 1:} First notice that for any $r>0$, there exists an integer $N_0(r) := \lceil 4r^2/\delta^2\rceil >0$ such that $\theta^* + h/\sqrt{n} \in B_{\theta^*}(\delta)$ whenever $h \in \overline B_{\bf{0}}(r)$ and $n \ge N_0(r)$. To see that this is true, notice that if $||h|| \le r$ and $n \geq 4r^2/\delta^2$, then $||h||/\sqrt{n} \le \delta/2< \delta$. This will ensure that the function $f_n(g, h)$ is well-defined whenever $g, h \in \overline{B}_{\bf{0}}(r)$.

Recall the definition of the $\alpha$-posterior $\pi_{n,\alpha}$ in \eqref{eqn:alpha-posterior}, as well as that of the scaled densities  
$\pi_{n,\alpha}^{LAN}(h\,|\,X^n)=n^{-1/2}\pi_{n,\alpha}(\theta^*+h/\sqrt{n}\,|\,X^n)$ 
and $\pi_n(h)=n^{-1/2}\pi(\theta^*+h/\sqrt{n})$ introduced in the proof of Theorem \ref{thm:BVMT-alpha}. Then we see that for any two sequences $\{h_n\},\{g_n\}$ in $\overline B_{\bf{0}}(r)$ and $n>N_0(r)$, 
\begin{align}
 \nonumber \frac{\pi_{n,\alpha}^{LAN}(g_n\,|\,X^n)}{\pi_{n,\alpha}^{LAN}(h_n\,|\,X^n)} = \frac{ \pi_{n,\alpha}\left(\theta^*+ \frac{g_n}{\sqrt{n}}\: \Big | \: X^n\right)}{ \pi_{n,\alpha}\left(\theta^*+ \frac{h_n}{\sqrt{n}}\: \Big | \: X^n\right)} &=   \frac{\left [ f_n\left(X^n \: \Big | \: \theta^*+ \frac{g_n}{\sqrt{n}}\right) \right] ^\alpha \pi\left(\theta^*+ \frac{g_n}{\sqrt{n}}\right)}{ \left [ f_n\left(X^n \: \Big | \: \theta^*+ \frac{h_n}{\sqrt{n}}\right) \right] ^\alpha \pi\left(\theta^*+ \frac{h_n}{\sqrt{n}}\right)  } \nonumber \\
  &=   \frac{\left [ f_n\left(X^n \: \Big | \: \theta^*+ \frac{g_n}{\sqrt{n}}\right) \right] ^\alpha \pi_n\left(g_n\right)}{\left [ f_n\left(X^n \: \Big | \: \theta^*+ \frac{h_n}{\sqrt{n}}\right) \right] ^\alpha \pi_n\left(h_n\right) } \, .\label{eq:posterior_LAN_ratio}
\end{align}
Thus by the definition in \eqref{eq:fn_theorem_def}, with the notation $s_n(h_n) = [f_n(X^n\,|\, \theta^* + h_n/\sqrt{n})/f_n(X^n\,|\, \theta^*)]^{\alpha}$,
$$f_n(g_n,h_n) = \Big \{1 - \frac{\phi_n(h_n) }{\pi_{n,\alpha}^{LAN}(h_n\:| \:X^n)} \, \frac{\pi_{n,\alpha}^{LAN}(g_n\:| \:X^n)}{\phi_n(g_n)} \Big \}^+ = \Big \{1 - \frac{\phi_n(h_n) s_n(g_n) \pi_n(g_n) }{\phi_n(g_n) s_n(h_n)  \pi_n(h_n)} \Big \}^+\, .$$
Recall that $ \phi_n(h_n)=  \phi(h_n \,| \, \Delta_{n, \theta^*}, V_{\theta^*}^{-1}/\alpha)$ and notice that since $||h_n||/\sqrt{n} < \delta$ as discussed at the beginning of the step 1 proof, the above is well-defined (i.e.\ $\pi_{n,\alpha}^{LAN}(h_n\:| \:X^n)$ is positive).

Next, 
for any sequence $h_n \in \overline B_{\bf{0}}(r)$, Assumption 1 implies 
\begin{equation}
    \log (s_n(h_n)) =  \alpha \log \left ( \frac{f_n(X^n\,|\, \theta^*+ \frac{h_n}{\sqrt{n}})}{f_n(X^n\,|\, \theta^*)} \right) =   h_n^{\top} \alpha V_{\theta^*}\Delta_{n,\theta^*}  - \frac{1}{2} h_n^{\top} \alpha V_{\theta^*} h_n  + o_{f_{0,n}}(1) \, ,
\end{equation}
and algebra shows that the log-likelihood of the normal density $\phi_n$ can be written as 
$$\log \phi_n(h_n)=- \frac{p}{2}\log (2\pi) + \frac{1}{2} \log( \det( \alpha V_{\theta^*}))  - \frac{1}{2} (h_n - \Delta_{n,\theta^*})^{\top} \alpha V_{\theta^*} (h_n - \Delta_{n,\theta^*}) \, ,$$
hence
\begin{align*}
    \log \Big ( \frac{s_n(h_n)}{\phi_n(h_n)}  \Big )& =   o_{f_{0,n}}(1) + \frac{p}{2}\log (2\pi)- \frac{1}{2} \log( \det( \alpha V_{\theta^*})) + \frac{1}{2}  \Delta_{n,\theta^*}^{\top} \alpha V_{\theta^*}  \Delta_{n,\theta^*}
\end{align*}
Now, for any sequence $g_n \in \overline B_{\bf{0}}(r)$, define
\begin{equation}
    \label{eq:aux_function_b}
b_n(g_n,h_n) \equiv  \frac{\phi_n(h_n) s_n(g_n) \pi_n(g_n) }{\phi_n(g_n) s_n(h_n)  \pi_n(h_n)} \, .
\end{equation}
Then we conclude 
\begin{equation}
    \log(b_n(g_n,h_n)) = \log \Big ( \frac{\phi_n(h_n) s_n(g_n) \pi_n(g_n) }{\phi_n(g_n) s_n(h_n)  \pi_n(h_n)}  \Big ) = o_{f_{0,n}}(1)\, ,
    \label{eq:oterm}
\end{equation} 
where we have used that $\pi_n(g_n), \pi_n(h_n) \to \pi(\theta^*)$ as $n \to \infty$.

Since $h_n, g_n$ are arbitrary sequences in $ B_{\bf{0}}(r)$, the result in \eqref{eq:oterm} is equivalent to saying that for any fixed $r$, there exists an integer 
$\widetilde{N}_0(r, \epsilon, \eta)$, 
such that for $n>\max\{\widetilde{N}_0(r, \epsilon, \eta), N_0(r)\}$:  
\begin{equation} 
P_{f_{0,n}}\left(\sup_{g_n, h_n \, \in \, \overline B_{\bf{0}}(r)}\left \lvert\log\left(b_n\left(g_n,h_n\right)\right)\right\lvert > \eta\right) \leq \epsilon \,.
    \label{eq:oterm2}
\end{equation} 

Next, notice that 
\begin{align*}
    |\log(b_n(g_n,h_n))|& \geq |\log(\min\{1, b_n(g_n,h_n)\})| = |\log(1 - f_n(g_n,h_n))| \geq f_n(g_n,h_n) \, ,
\end{align*}
where the equality follows since $f_n(g,h) = 1 - \min\{b_n(g,h), 1\}$ by definition, and the final inequality follows by noting that the function $f(x) = |\log(1-x)| - x$ is increasing for $x \in [0,1]$ and $f(0)=0$.
Thus, by \eqref{eq:oterm2},
for all $n > \max\{\widetilde{N}_0(r, \eta, \epsilon),  N_0(r)\}$,
\begin{equation} 
\begin{split}
P_{f_{0,n}}\left( \sup_{g, h \, \in \, \overline B_{\bf{0}}(r)}  f_n(g,h) > \eta \right)  & \leq P_{f_{0,n}}\left( \sup_{g_n, h_n \, \in \, \overline B_{\bf{0}}(r)}  f_n(g,h) > \eta \right) \\
&\le P_{f_{0,n}}\left(\sup_{g_n, h_n \, \in \, \overline{B}_{\bf{0}}(r)} |\log(b_n(g_n,h_n))| > \eta\right) \le \epsilon\, .
    \label{eq:oterm3}
    \end{split}
\end{equation} 
\noindent \textbf{Step 2:} 
Let $N^*(\epsilon, \eta) = \max\{\widetilde{N}_0(1, \epsilon, \eta), 4/\delta^2\}$ for $\widetilde{N}_0(r, \epsilon, \eta)$ defined just above \eqref{eq:oterm2} and for all $n > N^*(\epsilon, \eta)$ let
\[r_n = \max \{\, r \in \mathbb{R} \, | \, r \leq \delta\sqrt{n}/2  \text{ and }  n > \widetilde{N}_0(r, \epsilon, \eta)\,\} \, .\]
Notice this is well defined since for any $n > N^*(\epsilon, \eta)$ the choice $r=1$ is valid as $\frac{\delta}{2}(\sqrt{n}) > 1$ and $n > \widetilde{N}_0(1, \epsilon, \eta)$ by the definition of $N^*(\epsilon, \eta)$. For $n \leq N^*(\epsilon, \eta)$ we can define $r_n$ arbitrarily, e.g.\ $r_n = 1$.  First notice $r_n \rightarrow \infty$. Indeed,
\begin{align*}
r_n &= \max \{\, r \in \mathbb{R} \, | \, r \leq \delta\sqrt{n}/2  \text{ and }  n > \widetilde{N}_0(r, \epsilon, \eta)\, \} \leq \min\{  \delta\sqrt{n}/2 , \, \max \{ r \in \mathbb{R} \, |\,   n > \widetilde{N}_0(r, \epsilon, \eta)\}   \} \, .
\end{align*}
Clearly $\delta\sqrt{n}/2 \rightarrow \infty$, so $r_n \rightarrow \infty$ if $\max \{r \in \mathbb{R} \,\, | \,\,   n > \widetilde{N}_0(r, \epsilon, \eta)\} \rightarrow \infty$. This is true, since if it were not, there must be some $r_{max}$ such that $ \widetilde{N}_0(r, \epsilon, \eta) = \infty$ for $r > r_{max}$.  However Assumption \ref{asn:A1} holds for any compact set $K \in \mathbb{R}^p$, meaning for any ball $\overline{B}_{\textbf{0}}(r)$ with $r < \infty$, hence $ \widetilde{N}_0(r, \epsilon, \eta) < \infty$ for any $r < \infty$.

Finally, following the proof in step $1$, we show that for all $n > N^*(\epsilon,\eta)$,
\begin{equation}
\label{eq:toshow} 
\mathbb{P}_{f_{0,n}}\Big ( \sup_{g,h \, \in \, \overline{B}_{\bf{0}}(r_n)} f_n(g,h) > \eta \Big ) \le \epsilon \, .
\end{equation} 
First, $\theta^* + h_n/\sqrt{n} \in B_{\theta^*}(\delta)$ whenever $h_n \in \overline{B}_{\textbf{0}}(r_n)$ since $||h_n||/\sqrt{n} \leq r_n/\sqrt{n} \leq \delta/2$. This guarantees $f_n(g_n, h_n)$ is well-defined, as discussed in step 1. Then for all $n > N^*(\epsilon, \eta)$, by the $r_n$ definition,  $n > \widetilde{N}_0(r_n, \epsilon, \eta)$, hence by the work in \eqref{eq:oterm2}-\eqref{eq:oterm3}, the bound in \eqref{eq:toshow} holds.

\begin{lemma}\label{lemma3} Suppose Assumptions \ref{asn:A1} and \ref{asn:A2} hold. Let $(\mu_n, \Sigma_n)$ be a sequence such that $(\sqrt{n}(\mu_n-\theta^*),n\Sigma_n)$ is bounded in $f_{0,n}$-probability. Then,
\begin{equation*}
    \mathcal{K}( \phi(\cdot \, | \,\mu_n, \, \Sigma_n  )   \: ||\:  \pi_{n,\alpha}(\cdot  \,| \, X^n) )=\mathcal{K} \left ( \phi(\cdot \, | \,\mu_n, \, \Sigma_n  )   \: ||\:  \phi(\cdot \, | \, \widehat{\theta}_{\textrm{ML\,-\,$\mathcal{F}_n$}}, \, V_{\theta^*}^{-1} / (\alpha n)  ) \right) + o_{f_{0,n}}(1) \, .
\end{equation*}
 
\end{lemma}
 
\noindent \textbf{Proof:}
Using the change of variables $h \equiv \sqrt{n} ( \theta - \theta^* )$ and reparametrizing $\overline \mu_n \equiv \sqrt{n}(\mu_n - \theta^*)$, we have that 
\begin{align}
\nonumber
    \mathcal{K}\left(\phi(\cdot \, | \,\mu_n, \, \Sigma_n  )\:||\:\pi_{n,\alpha}(\cdot  \,| \, X^n \right)) &= \int \phi(\theta \, | \, \mu_n, \, \Sigma_n  ) \log \left ( \frac{\phi(\theta \, | \, \mu_n, \, \Sigma_n  )}{\pi_{n,\alpha}( \theta \,| \, X^n)}   \right ) d\theta \\
   \nonumber &= \int \phi(h \, | \,\overline \mu_n, \, n\Sigma_n  ) \log \left ( \frac{n^{p/2}\phi(h \, | \,\overline \mu_n, \, n\Sigma_n  )}{\pi_{n,\alpha}(\theta^* + h/\sqrt{n} \,| \, X^n)}   \right ) dh \\
    &= I_1+I_2+I_3 \, ,
    \label{eq:lem3_KL}
\end{align}
where
\begin{align*}
I_1 \equiv &  \int \phi(h \, | \,\overline \mu_n, \, n\Sigma_n  )\: \log \phi(h \, | \,\overline \mu_n, \, n\Sigma_n  ) dh \, ,\\
I_2 \equiv & - \int \phi(h \, | \,\overline \mu_n, \, n\Sigma_n  )\:  \log \left ( \frac{\pi_{n,\alpha}(\theta^* + h/\sqrt{n}\,|\,X^n)}{\pi_{n,\alpha}(\theta^*\,|\,X^n)} \right ) dh \,, \\
I_3 \equiv & - \left [ \log \left (\frac{\pi_{n,\alpha}(\theta^*\,|\,X^n)}{\phi(\theta^* \, | \, \widehat{\theta}_{\textrm{ML\,-\,$\mathcal{F}_n$}}, \, V_{\theta^*}^{-1} / (\alpha n) )} \right ) + \log \left ( \frac{\phi(\theta^* \, | \, \widehat{\theta}_{\textrm{ML\,-\,$\mathcal{F}_n$}}, \, V_{\theta^*}^{-1} / (\alpha n)  )}{n^{p/2}} \right ) \right]   \, .
\end{align*}
 We will compute the three terms separately and show that their sum gives the desired result. The first term $I_1$ is the negative of the entropy of a Gaussian distribution with mean $\overline \mu_n$ and covariance-matrix $n \Sigma_n$. A direct computation of this quantity gives
\begin{equation}\label{eqn:I1-lemma3}
     I_1 = -\frac{p}{2} - \frac{p}{2}\log (2\pi) - \frac{1}{2} \log | n \Sigma_n| \, . 
\end{equation}
To compute the second term $I_2$, notice that
\[ \frac{\pi_{n,\alpha}(\theta^*+h/\sqrt{n}\,|\,X^n)}{\pi_{n,\alpha}(\theta^*\,|\,X^n)} = \frac{f_n(X^n\,|\,\theta^*+h/\sqrt{n})^{\alpha}\:\pi(\theta^*+h/\sqrt{n})}{f_n(X^n\,|\,\theta^*)^{\alpha}\: \pi(\theta^*)} \, ,\]
hence
\begin{equation}
\begin{split}
I_2 & = - \alpha \int \phi(h \, | \,\overline \mu_n, \, n\Sigma_n  )  \log \left( \frac{f_n(X^n\,|\,\theta^*+h/\sqrt{n})}{f_n(X^n\,|\,\theta^*)} \right) dh \\
&\qquad -\int \phi(h \, | \,\overline \mu_n, \, n\Sigma_n  )  \log \left( \frac{\pi(\theta^*+h/\sqrt{n})}{\pi(\theta^*)} \right)  dh \, ,
\label{eq:I2_split}
\end{split}
\end{equation}
and we consider the two terms on the right side of \eqref{eq:I2_split} separately. First, notice that by Assumption \ref{asn:A2} the second term is $o_{f_{0,n}}(1)$. For the first term, consider
Assumption \ref{asn:A1}, and we find
\begin{equation}
\begin{split}
& - \alpha \int \phi(h \, | \,\overline \mu_n, \, n\Sigma_n  )  \log \left( \frac{f_n(X^n\,|\,\theta^*+h/\sqrt{n})}{f_n(X^n\,|\,\theta^*)} \right) dh \\
&= -\alpha \int \phi(h \, | \,\overline \mu_n, \, n\Sigma_n  )\: \left( h^{\top} \, V_{\theta^*} \, \Delta_{n,\theta^*}  - \frac{1}{2} \, h^{\top} \, V_{\theta^*}\, h + R_n(h) \right)   dh \\
&= -\alpha \int \phi(h \, | \,\overline \mu_n, \, n\Sigma_n  )\: \left( h^{\top} \, V_{\theta^*} \, \Delta_{n,\theta^*}  - \frac{1}{2} \, h^{\top} \, V_{\theta^*}\, h  \right)   dh  +o_{f_{0,n}}(1) \, .
\label{eq:I2_T1}
\end{split}
\end{equation}
Plugging this into \eqref{eq:I2_split} and solving the integral, we find
\begin{equation}
\begin{split}
I_2 &= -\alpha\int \phi(h \, | \,\overline \mu_n, \, n\Sigma_n  )\:  \left( h^{\top} \, V_{\theta^*} \, \Delta_{n,\theta^*}  - \frac{1}{2} \, h^{\top} \, V_{\theta^*}\, h  \right)  dh+o_{f_{0,n}}(1) \\
& = - \alpha\, \overline \mu_n^{\top} \, V_{\theta^*} \, \Delta_{n,\theta^*} + \frac{\alpha}{2} \left( \overline \mu_n^{\top} \, V_{\theta^*} \,  \overline \mu_n + \tr(n \Sigma_n V_{\theta^*}) \right)+o_{f_{0,n}}(1) \\
&= \frac{\alpha}{2}\left( (\Delta_{n,\theta^*}-\overline\mu_n)^{\top} \, V_{\theta^*} \,  (\Delta_{n,\theta^*}-\overline \mu_n) -\Delta_{n,\theta^*}^{\top} \, V_{\theta^*} \, \Delta_{n,\theta^*} + \tr(n \Sigma_n V_{\theta^*})  \right)+o_{f_{0,n}}(1)\, .
\label{eq:I2_split2}
\end{split}
\end{equation}
Next, replacing $\Delta_{n,\theta^*} = \sqrt{n}(\widehat{\theta}_{\textrm{ML\,-\,$\mathcal{F}_n$}}-\theta^*)$ and $\overline \mu_n = \sqrt{n}(\mu_n - \theta^*)$  in \eqref{eq:I2_split2} yields
\begin{align}
\nonumber      I_2=& \: \frac{1}{2} (\widehat{\theta}_{\textrm{ML\,-\,$\mathcal{F}_n$}} - \mu_n)^\top (\alpha nV_{\theta^*}) (\widehat{\theta}_{\textrm{ML\,-\,$\mathcal{F}_n$}} - \mu_n) + \frac{1}{2}\tr(\Sigma_n \alpha n V_{\theta^*}) \\
    &  \quad- \frac{1}{2} (\widehat{\theta}_{\textrm{ML\,-\,$\mathcal{F}_n$}} - \theta^*)^\top (\alpha nV_{\theta^*})(\widehat{\theta}_{\textrm{ML\,-\,$\mathcal{F}_n$}}-\theta^*)  +o_{f_{0,n}}(1) \, .
    \label{eqn:I2-lemma3}
\end{align}
Let's now turn to the term $I_3$. We first claim that
\begin{equation}\label{eqn:aux0_lemma3}
   \log \left (\frac{\pi_{n,\alpha}(\theta^*\,|\,X^n)}{\phi(\theta^* \, | \, \widehat{\theta}_{\textrm{ML\,-\,$\mathcal{F}_n$}}, \, V_{\theta^*}^{-1} / (\alpha n) )} \right ) = o_{f_{0,n}}(1) \, ,
\end{equation}
and we will prove this in what follows.
Result \eqref{eqn:aux0_lemma3} implies that
\begin{align}
\nonumber
I_3 &=   - \log \left ( n^{-p/2} \phi(\theta^* \, | \, \widehat{\theta}_{\textrm{ML\,-\,$\mathcal{F}_n$}}, \, V_{\theta^*}^{-1} / (\alpha n) ) \right )+ o_{f_{0,n}}(1) \\
&=  \frac{p}{2}\log (2\pi) - \frac{1}{2} \log |\alpha V| +  \frac{1}{2} (\widehat{\theta}_{\textrm{ML\,-\,$\mathcal{F}_n$}} - \theta^*)^\top(\alpha n V_{\theta^*}) (\widehat{\theta}_{\textrm{ML\,-\,$\mathcal{F}_n$}} - \theta^*)+o_{f_{0,n}}(1) \, .\label{eqn:I3-lemma3}
\end{align}

Finally,  combining \eqref{eq:lem3_KL}, \eqref{eqn:I2-lemma3} and \eqref{eqn:I3-lemma3} we conclude that
\begin{equation*}
    \begin{split}
        & \mathcal{K}(\phi(\cdot \, | \,\mu_n, \, \Sigma_n  ) \:||\:\pi_{n,\alpha}(\cdot  \,| \, X^n )) \\
         &= -\frac{1}{2}\bigg[ \log | \alpha n V_{\theta^*}|+ \log |\Sigma_n | -\tr(\Sigma_n \alpha n V_{\theta^*})  - (\widehat{\theta}_{\textrm{ML\,-\,$\mathcal{F}_n$}} - \mu_n)^\top (\alpha nV_{\theta^*}) (\widehat{\theta}_{\textrm{ML\,-\,$\mathcal{F}_n$}} - \mu_n) +p\bigg] + o_{f_{0,n}}(1)\\
         &=\mathcal{K} \left ( \phi(\cdot \, | \,\mu_n, \, \Sigma_n  )   \: ||\:  \phi(\cdot \, | \, \widehat{\theta}_{\textrm{ML\,-\,$\mathcal{F}_n$}}, \, V_{\theta^*}^{-1} / (\alpha n)  ) \right) + o_{f_{0,n}}(1)\, ,
    \end{split}
\end{equation*}
 where the last equality follows from \eqref{eq:KLnormals}.

It therefore remains to verify \eqref{eqn:aux0_lemma3} in order to complete the proof.  
We first
notice that the change of variables $h \equiv \sqrt{n} ( \theta - \theta^* )$ leads to  
\begin{equation}\label{eqn:aux_lemma3}
    \frac{\pi_{n,\alpha}(\theta^*\,|\,X^n)}{\phi(\theta^* \, | \, \widehat{\theta}_{\textrm{ML\,-\,$\mathcal{F}_n$}}, \, V_{\theta^*}^{-1} / (\alpha n) )} = \frac{\pi_{n,\alpha}^{LAN}(0\,|\,X^n)}{\phi_n(0   )}\, ,
\end{equation}
where $\phi_n(h) \equiv {n^{-p/2}}\phi(h\,|\, \Delta_{n,\theta^*},\: V_{\theta^*}^{-1}/\alpha)$ and $\pi_{n,\alpha}^{LAN}(h\:| \: X^n) \equiv n^{-p/2} \pi_{n,\alpha}(\theta^*+h/\sqrt{n}\:| \: X^n)$, are scaled versions of the densities $\pi_{n,\alpha}$ and $\phi$. Consider the function $b_n(g,h)$ defined as 
\begin{equation*}
b_n(g,h) \equiv \left( \frac{\pi_{n,\alpha}^{LAN}(h\,|\,X^n)}{\phi_n(h  ) } \right) \left(\frac{\phi_n(g )}{\pi_{n,\alpha}^{LAN}(g \,|\,X^n)} \right) \, ,
\end{equation*}
 and note that it is the same as \eqref{eq:aux_function_b}. Therefore,  from using \eqref{eq:oterm2}, we can guarantee the existence of $r_n$ such that for any $\eta >0$,
\[ \lim
P_{f_{0,n}}\Big(\sup_{g, h \, \in \, \overline B_{\bf{0}}(r_n)}|\log(b_n(g,h))| > \eta\Big) = 0 \, .\]
This implies that $\sup_{g,h \in \overline B_0(r_n)} b_n(g,h) = 1 + o_{f_{0,n}}(1)$.
Define $c_n = (\int_{\overline B_0(r_n)} \pi_{n,\alpha}^{LAN}(h\,|\,X^n) dh)^{-1}$ and $d_n = (\int_{\overline B_0(r_n)} \phi_n(h  ) dh)^{-1}$. Using this notation, we have that \eqref{eqn:aux_lemma3} is equal to
\begin{equation*}
    \frac{d_n}{c_n} \int_{\overline B_0(r_n)}  \left(\frac{\pi_{n,\alpha}^{LAN}(0\,|\,X^n)}{\phi_n(0   )} \right) \left(\frac{\phi_n(g  )}{\pi_{n,\alpha}^{LAN}(g \,|\,X^n)}\right)  \, c_n \, \pi_{n,\alpha}^{LAN}(g\,|\,X^n) \,  dg \, ,
\end{equation*} 
which is lower---by definition of $b_n$---than
\[ \frac{d_n}{c_n} \sup_{g,h \in \overline B_0(r_n)} b_n(g,h) = \frac{d_n}{c_n} \left( 1 + o_{f_{0,n}}(1)  \right ) \, . \]
By the concentration assumption of the $\alpha$-posterior \eqref{eqn:concentration} we have that   $c_n \to 1$ in $f_{0,n}$-probability. Furthermore, from  Lemma 5.2 in \cite{kleijn2012bernstein}, it follows that $d_n \to 1$ in $f_{0,n}$-probability. This implies that
\[   \frac{\pi_{n,\alpha}^{LAN}(0\,|\,X^n)}{\phi_n(0 \, | \, \Delta_{n,\theta^*}, \, V_{\theta^*}^{-1} / \alpha )} \le 1 +  o_{f_{0,n}}(1) \, .\]
In a similar way, we can conclude
\[\frac{\phi_n(0 \, | \, \Delta_{n,\theta^*}, \, V_{\theta^*}^{-1} / \alpha )}{\pi_{n,\alpha}^{LAN}(0\,|\,X^n)} \le 1 +  o_{f_{0,n}}(1) \, . \]
Using \eqref{eqn:aux_lemma3}, and the two inequalities above, we can conclude \eqref{eqn:aux0_lemma3}.

\begin{lemma}\label{lemma4}
Let $(\mu_{n,\alpha}, \Sigma_{n,\alpha})$ be the sequence defined in \eqref{eqn:alpha-mean-linref} and \eqref{eqn:alpha-variance-linref}. Denote by $\theta^*$ the (pseudo-) true parameter of the illustrative example in Section \ref{section:Example}. Then, the sequence $(\sqrt{n}(\mu_{n,\alpha}-\theta^*),\: n \Sigma_{n,\alpha})$ is bounded in $f_{0,n}$-probability. Moreover, if $\widehat{\theta}_{ML}$ is the maximum likelihood estimator of $\theta^*$, we have that $n(\mu_{n,\alpha}-\widehat{\theta}_{ML})$ is bounded in $f_{0,n}$-probability.
\end{lemma}

\noindent \textbf{Proof:}
Using \eqref{eqn:alpha-mean-linref}, we have that $\sqrt{n}(\mu_{n,\alpha}-\theta^*)$ is equal to
$$\left( \frac{1}{n} \sum_{i=1}^{n} W_i W_i^\top + \frac{1}{\alpha n } \Sigma_{\pi} \right)^{-1} \left( \frac{1}{\alpha \sqrt{n} } \Sigma_{\pi} ( \mu_\pi - \theta^*) + \frac{1}{n} \sum_{i=1}^{n} W_i W_i^\top \: \sqrt{n} (\widehat{\theta}_{ML}-\theta^*) \right) \, ,$$
where 
\[ \widehat{\theta}_{ML} = \left( \frac{1}{n} \sum_{i=1}^{n} W_i W_i^\top \right)^{-1} \frac{1}{n} \sum_{i=1}^{n} W_i Y_i \, . \]

Since $\theta^*$ is the (pseudo-) true parameter and $\widehat{\theta}_{ML}$ is the maximum likelihood estimator, we have that $\sqrt{n} (\widehat{\theta}_{ML}-\theta^*)$ converges in distribution to a multivariate normal distribution, and $ n^{-1} \sum_{i=1}^{n} W_i W_i^\top$ converges to $ \mathbb{E}[W_iW_i^\top]$ in $f_{0,n}$-probability. By Slutsky's theorem, we conclude $\sqrt{n}(\mu_{n,\alpha}-\theta^*)$ converges in distribution, which implies that $\sqrt{n}(\mu_{n,\alpha}-\theta^*)$ is bounded in  $f_{0,n}$-probability.

Using \eqref{eqn:alpha-variance-linref}, we have that $n\Sigma_{n,\alpha}$ is equal to
$$ n\Sigma_{n,\alpha} \equiv  \frac{\sigma^2_u}{ \alpha}\left( \frac{1}{n} \sum_{i=1}^{n} W_i W_i^\top + \frac{1}{\alpha n } \Sigma_{\pi} \right)^{-1} \, ,$$
and this converges in $f_{0,n}$-probability to 
$\frac{\sigma^2_u}{ \alpha}( \mathbb{E}[W_i W_i^\top])^{-1} = \frac{1}{\alpha} V_{\theta^*}^{-1}.$
Then, it follows that $n\Sigma_{n,\alpha}$ is bounded in  $f_{0,n}$-probability.

Finally, algebra shows that $n(\mu_{n,\alpha}-\widehat{\theta}_{ML})$ is equal to
$$ \left( \frac{1}{n} \sum_{i=1}^{n} W_i W_i^\top + \frac{1}{\alpha n } \Sigma_{\pi} \right)^{-1} \left( \frac{1}{\alpha} \Sigma_{\pi} ( \mu_\pi - \widehat{\theta}_{ML})  \right) \,,$$
which converge in $f_{0,n}$-probability to 
$( \mathbb{E}[W_iW_i^\top] )^{-1} ( \frac{1}{\alpha} \Sigma_{\pi} ( \mu_\pi - \widehat{\theta}_{ML})).$
This implies that $n(\mu_{n,\alpha}-\widehat{\theta}_{ML})$ is bounded in  $f_{0,n}$-probability.

\vskip 0.2in

\bibliography{references}

\end{document}